\documentclass[10pt,twocolumn,letterpaper]{article}

\usepackage[pagenumbers]{wacv} 

\usepackage{graphicx}
\usepackage{amsmath}
\usepackage{amssymb}
\usepackage{booktabs}

\usepackage[pagebackref,breaklinks,colorlinks]{hyperref}

\usepackage{graphicx}	
\usepackage{amsmath}	
\usepackage{amssymb}	
\usepackage{booktabs}
\usepackage{times}
\usepackage{microtype}
\usepackage{epsfig}
\usepackage{caption}
\usepackage{float}
\usepackage{placeins}
\usepackage{color, colortbl}
\usepackage{stfloats}
\usepackage{enumitem}
\usepackage{tabularx}
\usepackage{xstring}
\usepackage{multirow}
\usepackage{xspace}
\usepackage{url}
\usepackage{subcaption}
\usepackage{xcolor}
\usepackage[hang,flushmargin]{footmisc}
\usepackage{tikz}
\usepackage{pgfplots}
\usepackage{pgf}
\usepackage{overpic}
\usepackage[utf8]{inputenc}

% ...

\newcommand{\supp}{supplementary\xspace}

\newcommand{\R}[1]{{%
    \textbf{%
        \ifstrequal{#1}{1}{\textcolor{red}{R#1}}{%
        \ifstrequal{#1}{2}{\textcolor{blue}{R#1}}{%
        \ifstrequal{#1}{3}{\textcolor{magenta}{R#1}}{%
        \ifstrequal{#1}{4}{\textcolor{teal}{R#1}}{%
                           \textcolor{cyan}{R#1}%
        }}}}%
    }%
}}

\newcommand*{\inparagraph}[1]{\noindent\textbf{#1}\hspace{0.5em}}

\usepackage{booktabs} 
\usepackage{tabularx}
\usepackage{siunitx}
\usepackage{pifont}
\usepackage{amsmath}
\usepackage[nohyperlinks, printonlyused, withpage, smaller]{acronym}
\usepackage{comment}
\usepackage{svg}
\usepackage{tikz}
\usepackage{pgfplots}
\pgfplotsset{compat=newest}
\usepackage[dvipsnames]{xcolor}
\usetikzlibrary{calc}
\usepackage{minitoc} 
\usepackage{xr-hyper}  
\usepackage{tocloft}

\usepackage{xcolor} 
\definecolor{coral}{RGB}{255,94,77} 
\definecolor{softgreen}{RGB}{60,179,113} 
\definecolor{skyblue}{RGB}{60,179,113} 

\usepackage{caption}

\newcommand{\cbarm}[4]{%
  {\begingroup\color{#1}\rule{#2}{#3}\endgroup}%
  \hspace{0.5em}#4%
}

\usepackage[capitalize]{cleveref}
\crefname{section}{Sec.}{Secs.}
\Crefname{section}{Section}{Sections}
\Crefname{table}{Table}{Tables}
\crefname{table}{Tab.}{Tabs.}

\def\wacvPaperID{2186} 
\def\confName{WACV}
\def\confYear{2025}

\newif\ifreview 
\newif\ifarxiv 
\newif\ifcamera 
\newif\ifrebuttal 

\newcommand{\ourmethod}{\ac{IDEAL-M3D}\xspace}
\newcommand{\metricRequested}{NAURC}
\newcommand{\cmark}{\ding{51}}
\newcommand{\xmark}{\ding{55}}

\newcommand{\coreSetBox}{\textit{Core-Set Box}$_{\text{\textit{3D}}}$\xspace\xspace}

\def\paperTitle{IDEAL-M3D: Instance Diversity-Enriched
Active Learning\\[2mm] for Monocular 3D Detection}
\acrodef{RoI}[RoI]{region of interest}
\acrodef{SAMv2}[SAMv2]{Segment Anything Model 2}
\acrodef{AL}[AL]{Active Learning}
\acrodef{ERCS}[ERCS]{Espistemic Representation for Core-Set Sampling}
\acrodef{ACW}[ACW]{Adaptive Class Warmup}
\acrodef{IDEAL-M3D}[IDEAL-M3D]{Instance Diversity-Enriched Active Learning for Monocular 3D Detection}
\acrodef{NAURC}[NAURC]{Normalized Area under the Requested Curve}
\acrodef{AUC}[AUC]{Area under the Curve}
\acrodef{SOTA}[SOTA]{state-of-the-art}
\acrodef{M3D}[M3D]{Monocular 3D Detection}

\acrodef{AURC}[AURC]{Area under the Requested Curve}
\acrodef{PCA}[PCA]{principal component analysis}
\acrodef{MI}[MI]{mutual information}

\title{\paperTitle}

\author{
Johannes Meier$^{1,2,3,4,\dagger,}$\footnotemark[1]  \and Florian Günther $^{3,\dagger,}$ \and Riccardo Marin $^{3,4}$ \and Oussema Dhaouadi $^{1,3,4}$ \and Jacques Kaiser $^{1}$ \and Daniel Cremers $^{3,4}$
}

\begin{document}
\twocolumn[{%
\maketitle
\vspace{-1cm}
\begin{center}
\large	
    $^{1}$ DeepScenario\qquad
    $^{2}$ ETH Zurich\qquad
    $^{3}$ TU Munich\qquad
    $^{4}$ MCML 
\end{center}
\vspace{1cm}
	\renewcommand\twocolumn[1][]{#1}%
	\maketitle
	\vspace{-2.6em}
    \includegraphics[width=0.68\linewidth]{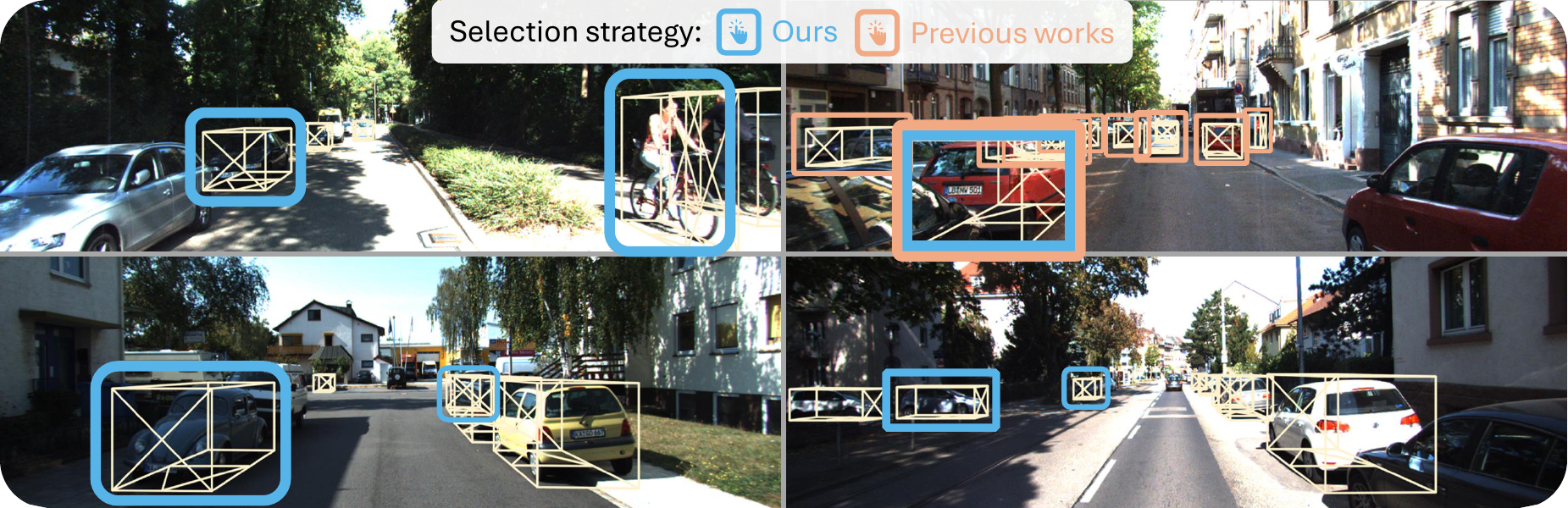}
    \includegraphics[width=0.31\linewidth]{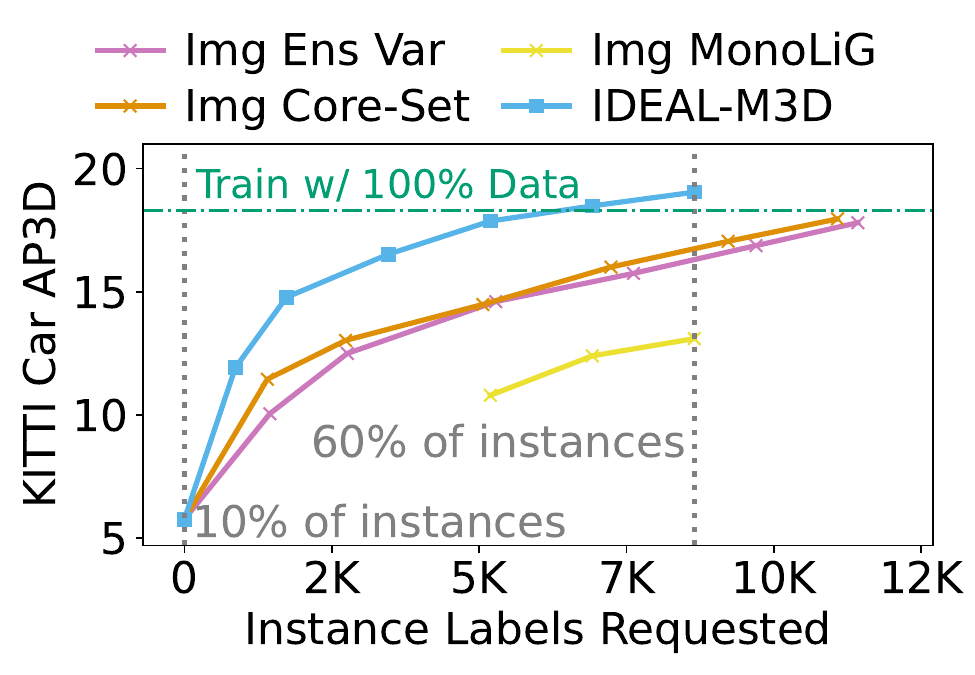}
    \vspace{-0.2cm}
    
    \captionof{figure}{\acs{IDEAL-M3D} is the first instance-based active learning method for monocular 3D detection. \textbf{Left}: While previous active learning approaches select entire images for labeling, we identify the most informative object instances (difference is highlighted in rounded boxes). \textbf{Right}: Our approach achieves full supervised performance using only 50-60\% of the labeled boxes, significantly outperforming existing active learning methods across the majority of object categories (KITTI \cite{kitti} validation set, $AP_{3D|R_{40}}^{0.7} Mod$.).}
	\label{fig:teaser}
	\vspace{0.85em}
 
}]
\footnotetext[1]{j.meier@tum.de\space\space\space $^\dagger$ Equal contribution} 
\begin{abstract}
Monocular 3D detection relies on just a single camera and is therefore easy to deploy. Yet, achieving reliable 3D understanding from monocular images requires substantial annotation, and 3D labels are especially costly. To maximize performance under constrained labeling budgets, it is essential to prioritize annotating samples expected to deliver the largest performance gains. This prioritization is the focus of active learning. Curiously, we observed two significant limitations in active learning algorithms for 3D monocular object detection. 
First, previous approaches select entire images, which is inefficient, as non-informative instances contained in the same image also need to be labeled. Secondly, existing methods rely on uncertainty-based selection, which in monocular 3D object detection creates a bias toward depth ambiguity. Consequently, distant objects are selected, while nearby objects are overlooked.

To address these limitations, we propose  IDEAL-M3D, the first instance-level pipeline for monocular 3D detection.
For the first time, we demonstrate that an explicitly diverse, fast-to-train ensemble improves diversity-driven active learning for monocular 3D. We induce diversity with heterogeneous backbones and task-agnostic features, loss weight perturbation, and time-dependent bagging. IDEAL-M3D shows superior performance and significant resource savings: with just 60\% of the annotations, we achieve similar or better AP$_{3D}$ on KITTI validation and test set results compared to training the same detector on the whole dataset. 
\end{abstract}

\section{Introduction}
\label{sec:intro }

3D object detection estimates 3D bounding boxes of objects visible in 2D images by predicting their position, orientation, and dimensions. Identifying objects and their geometrical properties is a fundamental step toward understanding the environment, and with the progress in the autonomous driving industry ~\cite{progress_driving,progress_driving2}, it has gained significant traction. The monocular setting is particularly compelling, since relying on a single RGB camera is flexible, cheap, and simple to set up. Although estimating 3D properties such as depth from a monocular view is theoretically impossible, as discussed in ~\cite{monoground,monodde,mono_cd,monodle,deviant}, strong data priors can address most of these ambiguities in practice. This leads to reliable and practically useful predictions ~\cite{monolig}.

As a consequence, annotated data has become a pivotal element in fostering research in the field, and the research community has put tremendous effort into curating large-scale datasets ~\cite{survey_recent_deep_al,Waymo,rope3d,roscenes}. However, 3D data annotations are labor-intensive, costly, and difficult to scale. To limit such effort, \ac{AL} offers a compelling alternative by identifying the most informative unlabeled samples to annotate. Such selection reduces the redundant annotations ~\cite{survey_recent_deep_al,comp_al} and is particularly valuable for \ac{M3D}, where unlabeled data is abundant ~\cite{rope3d,roscenes}. 


Existing \ac{AL} methods for \ac{M3D}, such as Efficient AL \cite{efficient_al} and MonoLiG \cite{monolig}, operate at the image level. They request entire images rather than specific objects. This induces unnecessary annotation overhead because every object in a selected image must be labeled, including trivial or already well-modeled instances (see \cref{fig:teaser}). It also obscures the true annotation cost, which scales with the number of instances to be annotated \cite{ComPAS}. As shown by Lyu et al.~\cite{ComPAS}, labeling capacity is better utilized when annotating instances rather than images. In addition, existing \ac{AL} methods for \ac{M3D} \cite{monolig,efficient_al} are primarily uncertainty-based and rank samples by low predictive confidence or high disagreement (\eg, ensemble variance). At the instance level, such criteria over-select distant objects, as they exhibit high aleatoric and epistemic uncertainty but deliver limited accuracy gains. We hypothesize that prior work \cite{monolig,efficient_al} overlooked this bias because image-level selection compels annotating all objects in each chosen image. This mixes informative and uninformative instances and masks the inefficiency.

We revisit \ac{AL} for \ac{M3D} from a diversity perspective. First, we formalize an instance-based \ac{AL} pipeline and make the instance the unit of annotation. Then, we introduce \textbf{\acf{IDEAL-M3D}}: The first instance-based \ac{AL} method for \ac{M3D} (\cf \cref{tab:comparison}). A key limitation of prior diversity-based approaches is that they rely on features from a single detector, which introduces model- and task-specific biases. Motivated by information theory, our core idea is to use ensembles for diversity-based selection. We estimate each instance’s diversity jointly over a heterogeneous ensemble and combine complementary representations to reduce single-model bias. While ensembles are common for uncertainty estimation, to the best of our knowledge this is the first use of ensembles to drive diversity-based \ac{AL}.

Our contributions include several measures to increase the heterogeneity of the ensemble. This involves random loss weight sampling, data sampling and the utilization of different backbones. As an additional benefit, these choices cut the additional training time by over 50\% compared to a vanilla ensemble. Furthermore, we add visual features from a pre-trained image-based autoencoder to the ensemble. This adds no extra detector-training time and lets the selection exploit both task-related and task-independent cues. 

We validate our approach on KITTI \cite{kitti}, Waymo \cite{Waymo}, and additionally on Rope3D \cite{rope3d} for cross-perspective robustness. \mbox{IDEAL-M3D} achieves state-of-the-art results across all datasets. On the KITTI \cite{kitti} validation and test set, it reaches 100\% of the full-data performance using only 60\% of the labels. In some cases, it even surpasses the fully supervised counterpart. On Rope3D\cite{rope3d} \ourmethod achieves over 97\% accuracy with just 25\% of data, while on Waymo over 95\% with just 25\% of data. Finally, we propose \textbf{\acf{NAURC}}, a novel \ac{AL} metric that allows for direct comparison across instance- and image-based methods. The contributions of this work are:
\begin{itemize}
    \item We design the first instance-based \ac{AL} pipeline for \ac{M3D} and show that uncertainty-based selection underperforms at the instance level, where it over-selects distant objects with limited benefit.
    \item We introduce \ac{IDEAL-M3D}, a diversity-based selector with highly diverse ensemble features that reduces training overhead by over 50\% compared to the vanilla ensemble. Furthermore, we show that it leads to a complementary performance gain when these task-related detector features are combined with task-independent visual features.
    \item Our approach is simple to implement yet achieves state-of-the-art performance on KITTI \cite{kitti}, Waymo \cite{Waymo}, and Rope3D \cite{rope3d}.
\end{itemize}

\begin{table}
\small
\caption{\label{tab:prevworks} \textbf{Comparative analysis of key related research and methodologies.} Our \ac{AL} approach is the first instance-based method for \acf{M3D}.}
\vspace{-0.6em}
\begin{tabular}{@{}lcccc@{}}
\toprule
\multirow{2}{*}{Methods} & \multirow{2}{*}{Boxes} & \multirow{2}{*}{Modality} & Instance- & Diversity-\\
& & & based & based \\ \midrule
ComPAS ~\cite{ComPAS} & 2D    & Image & \cmark & \xmark  \\
QBox ~\cite{q_box} & 2D      & Image & \cmark & \xmark  \\ 
DDFH ~\cite{ddfh} & 3D & LiDAR & \xmark & \cmark \\
Efficient AL ~\cite{efficient_al} & 3D & Image & \xmark & \cmark \\
\multirow{2}{*}{MonoLiG ~\cite{monolig}} & \multirow{2}{*}{3D} & Image \& & \multirow{2}{*}{\xmark} & \multirow{2}{*}{\xmark} \\
& & LiDAR \\ 
\midrule
\textbf{Ours} & \textbf{3D} & \textbf{Image} & \cmark & \cmark \\ \bottomrule 
\end{tabular}
\vspace{-1.2em}
\label{tab:comparison}
\end{table}


\section{Related Work}
\label{sec:related }

\subsection{Monocular 3D Object Detection (M3D)}
\acf{M3D} aims to recover 3D information of objects from single 2D images ~\cite{mono_cd,monoflex,monodde,gupnet}. The research community has made a sustained effort to curate datasets and benchmarks for it, such as  KITTI ~\cite{kitti}, Waymo ~\cite{Waymo}, CDrone ~\cite{groundmix}, Rope3D ~\cite{rope3d} and V2X-I ~\cite{v2x,v2x-seq}. Such data often comes with additional LiDAR information, useful for providing additional depth supervision during training (e.g., RD3D ~\cite{rd3d}, CaDDN ~\cite{caddn}, MonoNeRD ~\cite{mononerd}, OccupancyM3D ~\cite{occupancym3d}, MonoTAKD \cite{monotakd}). However, LiDAR is not always available \cite{dsc3d,roscenes}, limiting the scalability of the approach. Methods that rely solely on 3D bounding box annotations such as MonoCon ~\cite{monocon}, MonoDETR ~\cite{monodetr}, MonoCD ~\cite{mono_cd}, MonoMAE ~\cite{monomae} and MonoUNI ~\cite{monouni} are compelling since they use only RGB information. To compensate for missing 3D information, these methods require a large number of annotated images. Among the most recent approaches, MonoDiff ~\cite{monodiff} frames the task as a denoising diffusion process, and MonoLSS ~\cite{monolss} employs Gumbel Softmax to identify depth-relevant features. MonoLSS has particularly strong performance and independence from LiDAR supervision. 
We primarily evaluate on the standard benchmarks, KITTI ~\cite{kitti} and Waymo ~\cite{Waymo}, and additionally include Rope3D ~\cite{rope3d} to assess generalization to traffic-view scenarios.

\subsection{Active Learning}
\ac{AL} reduces annotation costs by selecting the most informative samples for labeling. Traditional \ac{AL} methods can be divided into uncertainty-based approaches (e.g., Maximum Entropy ~\cite{entropy}, BALD ~\cite{bald}), diversity-based approaches (e.g., Core-Set ~\cite{core_set}, DiscAL ~\cite{disc_al}), and hybrid methods ~\cite{uwe,comp_al}. For an exhaustive survey on the topic, we point to ~\cite{comp_al}.
However, these methods are primarily designed for classification tasks, whereas \ac{M3D} is a regression problem.

\inparagraph{Active Learning for 2D Object Detection.} 
PPAL ~\cite{plug_play} combines weighted entropy scores with diversity filtering, while MI-AOD ~\cite{mi-aod} first performs adversarial training and later measures uncertainty via image-instance inconsistencies. During our investigation, we found that uncertainty-based methods are not suitable for instance-based \ac{AL} in 3D object detection since they tend to be biased toward the farthest objects in the scene.

\inparagraph{Active Learning for 3D Object Detection.}
In LiDAR-based 3D detection, \ac{AL} methods such as STONE ~\cite{stone} and CRB ~\cite{crb} identify representative prototypes in gradient space. 
STONE ~\cite{stone} incorporates uncertainty estimation via Monte Carlo dropout, while CRB ~\cite{crb} optimizes for uniform point density. 
DDFH ~\cite{ddfh} compresses high-dimensional features and bounding box information using t-SNE ~\cite{tsne} and employs Gaussian mixture models for diverse sample selection. 
KECOR ~\cite{kecor} leverages neural tangent kernels to quantify sample uniqueness. 
These methods address LiDAR-specific challenges such as point density diversity ~\cite{ddfh,crb}. In contrast, monocular image-based detection requires different handling strategies as it confronts different challenges, like the estimation of depth from a single image. 

For \ac{M3D}, \ac{AL} remains underexplored. 
Efficient Active Learning ~\cite{efficient_al} estimates epistemic uncertainty using heatmaps and maximizes diversity using a 2D detector trained on MS COCO ~\cite{ms-coco} but fails to integrate highly task-related features. 
MonoLiG~\cite{monolig} is an uncertainty-based approach that relies on a LiDAR-trained teacher and an ensemble of five student models. 
This design imposes high training costs, yet it lacks mechanisms to reduce the ensemble overhead or to explicitly encourage ensemble diversity. In our experiments, it still underperforms our approach. 
In contrast, we eliminate LiDAR dependency and adopt a diversity-driven selection tailored to \ac{M3D}.

\inparagraph{Instance-based Active Learning.}
All mentioned \ac{AL} methods select entire scenes or images. This requires annotating all the objects in the image, even when they are not relevant. On the contrary, instance-based approaches aim to identify the individual objects which are the most informative to annotate. Didari et al. ~\cite{bayesian_al} and ViewAL ~\cite{view_al} design specific techniques for segmentation by selecting pixels or superpixels in the images. 
In 2D object detection, methods like ComPAS ~\cite{ComPAS} and QBox ~\cite{q_box} rely on uncertainty, for instance selection, which, as already mentioned, is suboptimal for \ac{M3D}. Instead, we use a representative-based selection.

\section{IDEAL-M3D}

\begin{figure*}
  \centering
  \includegraphics[width=0.99\textwidth]{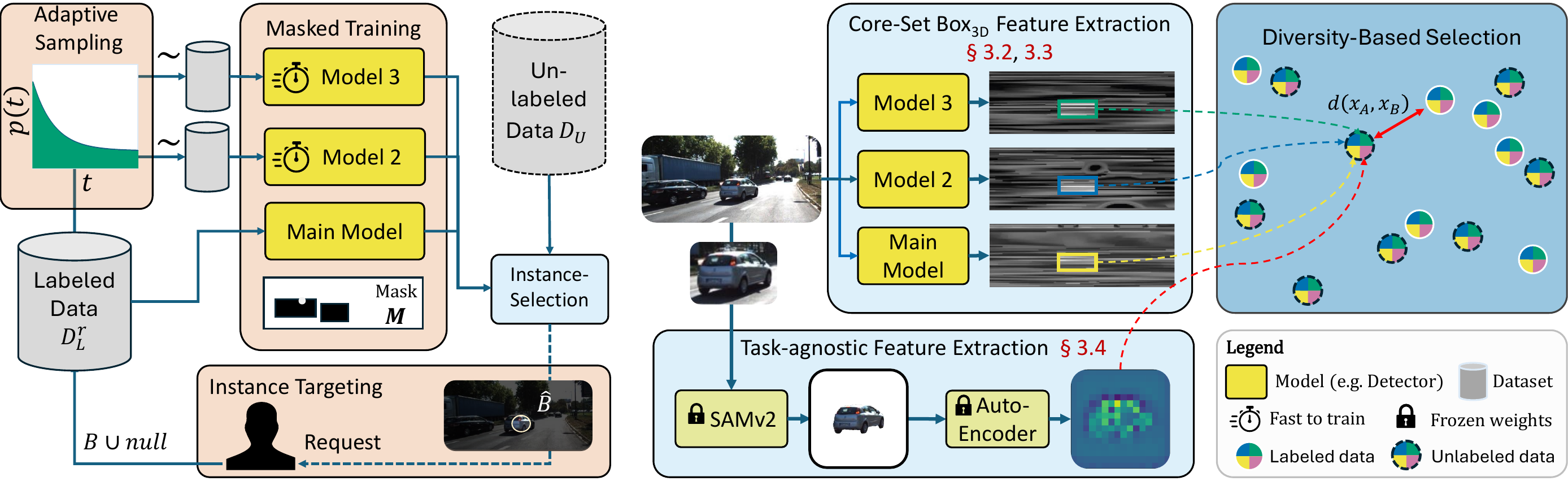}
  \vspace{-0.26cm}
  \caption{
  \textbf{Overview of IDEAL-M3D.} \textbf{Left:} Our instance-based \ac{AL} pipeline couples precise instance targeting with time-adaptive sampling, minimizing expert effort while remaining training-time efficient (\cref{sec:04_method_instance_based}). 
  \textbf{Right:} We maximize feature-space coverage by fusing Core-Set selection with an explicitly diverse, fast-to-train ensemble and task-agnostic visual embeddings, yielding robust geometry-aware selection under modest compute (\cref{sec:coreset_baseline,sec:diverse_ensembles,sec:task_agnostic_features}). 
  IDEAL-M3D uniquely integrates diversity-based selection with an ensemble purpose-built for representational diversity in \ac{M3D}, delivering label efficiency without the cost of conventional ensembles.
  }
  \label{fig:04_method_overview}
  \vspace{-1.4em}
\end{figure*}

We provide a high-level overview of \ac{IDEAL-M3D} (\cref{fig:04_method_overview}). In \cref{sec:04_method_instance_based}, we present our adaptation of the \ac{AL} pipeline to instance-based selection. The subsequent sections detail our instance selection strategy: Sec.~\ref{sec:coreset_baseline} adapts our diversity-based baseline Core-Set \cite{core_set} to \ac{M3D}, while Sec.~\ref{sec:diverse_ensembles} and Sec.~\ref{sec:task_agnostic_features} show how diverse ensembles and task-agnostic features synergistically broaden feature-space coverage for more effective selection.

\subsection{Instance-Based AL for M3D}
\label{sec:04_method_instance_based}

\inparagraph{Initialization.} We begin by labeling a small random subset of images to initialize the detector. Then we train our model over multiple \ac{AL} rounds. In each round, the network runs inference on all unlabeled images. We then use the predictions (in particular, the 2D bounding box center) to propose candidates for labeling until the labeling budget of the current round is exhausted. 

\inparagraph{Processing of Unlabeled Instances.} During selection, 3D box locations are an unreliable matching criterion as depth can be very imprecise, especially at early iterations or at high distances. Instead, we identify the object to annotate as the one whose 2D center is closest to the requested 2D center. To avoid requiring the annotator to scan the entire image, we restrict the search to a depth-dependent window of size $(r_x, r_y)$ around the predicted center; see the \supp for the precise definition of this radius. We then select instances to label in the amounting to $5$–$10\%$ of the global budget per round. This selection lies at the core of our approach and is detailed in the next section.

\inparagraph{Training.} After selecting training instances, we update the model with the loss used by the baseline detector (MonoLSS~\cite{monolss}, MonoCon~\cite{monocon}). Because only a small set of instances is labeled, the classification head tends to assign unlabeled objects to the background, penalizing rare classes. Annotating all objects per image (as in image-based \ac{AL}) mitigates this but is costly; in instance-based \ac{AL}, rare classes may be missed.

Hence, we introduce a binary mask $M_{i,j}$ for the classification loss to ignore uncertain regions. Let the predicted probability at pixel $(i,j)$ for class $c$ be $\hat{p}_{c,i,j}$ and the ground-truth be $p_{c,i,j}$. We define the binary Mask $M$ as:
\begin{equation}
    M_{i,j} =
    \begin{cases}
    0 & \text{if } \sum_c p_{c,i,j}=0 \text{ or } (i,j) \text{ falls inside the}\\
      & \text{2D box of a previously predicted object},\\
      & \text{that is still unlabeled}\\
    1 & \text{otherwise}.
    \end{cases}
\end{equation}
We compute the masked objectness loss, where $\ell$ denotes the weighted focal loss~\cite{centernet}, as:
\begin{equation}
    L_{cls}^{\text{masked}} = \sum_{c} \sum_{i,j} M_{i,j} \cdot \ell(\hat{p}_{c,i,j}, p_{c,i,j}).
\end{equation} 

\subsection{Core-Set Box$_{3D}$}
\label{sec:coreset_baseline}

In our instance-based regime, popular uncertainty-based methods underperform (\cf \cref{tab:main_results}): they often select far-away objects with high epistemic or aleatoric uncertainty, which are less useful for \ac{M3D} than closer, well-resolved instances (\cf \cref{sec:div_vs_unc_methods}). These observations motivate a diversity-based baseline. We find that Core-Set~\cite{core_set}, a simple diversity-based approach, performs surprisingly well in this setting, so we adopt it as our baseline.

Core-Set selects the unlabeled instance farthest from the labeled set in feature space. At labeling round $r$, with features $f(x)$, labeled set $\mathcal{D}_L^r$, and unlabeled set $\mathcal{D}_U^r$, it scores an unlabeled instance $x$ by
\begin{equation}
    \mathrm{score}(x) \;=\; \min_{z\in \mathcal{D}_L^r} \| f(x)-f(z) \|_2,
\end{equation}
and choose the next query
\begin{equation}
    x^* \;=\; \arg\max_{x\in \mathcal{D}_U^r}\; \mathrm{score}(x).
\end{equation}
This greedy step repeats until the round budget is exhausted.

A first weakness of the classic variant is that $f(x)$ is a penultimate classification feature. This is suboptimal for \ac{M3D} because they are trained to be invariant to geometric factors (size, depth, pose) that dominate 3D detection quality. We therefore extract features immediately before the 3D detection heads, flatten them, and use these as $f(x)$ in the Core-Set distance; we refer to this instance-level adaptation as \coreSetBox. These pre-head features retain the 3D cues required for 3D box prediction, thereby aligning the distance metric with the information actually used for \ac{M3D}.

\subsection{Diverse ensembles}
\label{sec:diverse_ensembles}

\inparagraph{Motivation.} One important limitation of Core-Set~\cite{core_set} (and inherited by \coreSetBox) is the reliance on a single representation $f(\cdot)$. A single model’s inductive biases and training dynamics (e.g., feature geometry shaped by loss composition, augmentations, and optimization) can dominate the notion of “distance,” risking that other task-relevant modes are overlooked. We remove this single-model bias by leveraging an ensemble. Prior work has used ensembles for uncertainty-based scoring; to our knowledge, we are the first to deploy an ensemble explicitly for diversity-based selection in \ac{M3D}. Concretely, we replace $f(\cdot)$ with $3$ detector-specific embeddings of the same instance $x$, denoted $f_1(x), f_2(x), f_3(x)$, and measure distances with a weighted cosine across views:
\begin{equation}
\begin{aligned}
    d(x_A, x_B)
    \;=\; \sum_{i \in \{1,2,3\}} \lambda_i \cdot d_{\cos}\!\big(f_i(x_A), f_i(x_B)\big),
\end{aligned}
\label{eq:distance}
\end{equation}
where $\lambda_i \ge 0$ and $d_{\cos}(u,v)= 1 - \tfrac{u^\top v}{\|u\|_2\,\|v\|_2}$. We then apply the Core-Set~\cite{core_set} selection in this fused space:
\begin{equation}
    D_{\mathrm{ens}}(x \mid \mathcal{D}_L^r) \;=\; \min_{z\in\mathcal{D}_L^r} d(x,z).
\end{equation}
\begin{equation}
    x^* \;=\; \arg\max_{x\in \mathcal{D}_U^r} D_{\mathrm{ens}}(x \mid \mathcal{D}_L^r).
\end{equation}

\inparagraph{Information Theory Intuition.} For intuition, we analyze the ensemble representation as a concatenation $F_{\mathrm{ens}}(x)=[f_1(x),f_2(x),f_3(x)]$ and omit $x$ for readability. By the chain rule of \ac{MI},
\begin{equation}
\label{eq:mi_ensemble_gain}
\begin{aligned}
    I(F_{\mathrm{ens}};\mathcal{D}_L^r)
    &= I(f_1;\mathcal{D}_L^r) + \sum_{m=2}^{3} I\!\big(f_m;\mathcal{D}_L^r \mid f_{<m}\big) \\
    &\ge I(f_1;\mathcal{D}_L^r).
\end{aligned}
\end{equation}
Because the concatenation order is arbitrary, we can permute the views so that any $f_m$ appears first, yielding:
\begin{equation}
\label{eq:mi_ensemble_gain2}
\begin{aligned}
    I(F_{\mathrm{ens}};\mathcal{D}_L^r) &\ge \max_{m\in\{1,2,3\}} I(f_m;\mathcal{D}_L^r).
\end{aligned}
\end{equation}

The result in \eqref{eq:mi_ensemble_gain} and \eqref{eq:mi_ensemble_gain2} gives strict gains when added views contribute a nonredundant signal.

\inparagraph{Efficiency.} Ensembles are often weakly diverse, since their members often differ mainly by random initialization and data-loading order. Also, they are slow to train. To address both issues, we train one main model in standard fashion (also used for evaluation) and add two auxiliary models that are faster and intentionally diversified. We introduce the following measures to increase representational diversity while reducing training time:

\begin{itemize}
    \item \textbf{Heterogeneous, smaller backbones.} For the auxiliary models, we choose different backbones to induce diversity via both distinct architectures and initializations from their respective pretrained weights. For fairness, all detectors (main and auxiliary) are initialized from ImageNet \cite{imagenet} pretraining. In our experiments, we use RepViT-M1.0~\cite{repvit} and MobileNet4-Conv-M~\cite{mobilenetv4}, selected for (i) reasonable accuracy and (ii) faster training than the main model (see \cref{fig:backbone_ablation}).
    \item \textbf{Shorter schedules.} Each auxiliary model trains for one third of the main model’s epochs, retaining most of the diversity signal while substantially reducing compute.
    \item \textbf{Multi-task loss weighting perturbations.} \ac{M3D} comprises several subtasks (2D box, 3D offset, dimensions, depth, orientation, confidence, classification). At the start of each round $r$, for each auxiliary model $m$ we sample per-subtask $\Psi$ loss multipliers $w^{m}_{r,\Psi} \sim \mathrm{Uniform}(1-\delta, 1+\delta)$ to induce mild specialization.
    \item \textbf{Time-dependent bagging.} Later rounds dominate training cost because the labeled set has grown, whereas early rounds are label-sparse and benefit from using a larger fraction of available labels; accordingly, we adopt a decreasing subsampling schedule over training progress. At training round $r$ with progress $t \in [0,1]$ (where $t \triangleq \tfrac{|\mathcal{D}_L^r|}{|\mathcal{D}_L^r| + |\mathcal{D}_U^r|}$), we subsample the labeled data with:
    \begin{equation}
        s(t) \;=\; 0.5 \;+\; 0.4\,\exp(-\alpha\, t), \qquad \alpha>0,
    \end{equation}
    which monotonically decreases the sampled fraction over time, using more data early and less later.
\end{itemize}

\subsection{Task-agnostic features}
\label{sec:task_agnostic_features}

Even with a multimodel ensemble, relying solely on task-specific features can bias the metric toward detector idiosyncrasies and under-represent appearance variation. To complement these signals, we augment the ensemble feature space with task-agnostic visual features that capture object appearance (e.g., texture, color, shape, local context).

Concretely, for each candidate instance we segment the object with SAMv2~\cite{sam} and encode the masked crop using a pretrained Stable Diffusion autoencoder~\cite{ldm} to obtain a compact flattened visual embedding $f_{\mathrm{vis}}(x)$. Since the SAM mask is only used for feature selection and not for the final selection itself, a slight imperfection in the mask is not critical. We then integrate these features into our cosine-based distance and select the highest-scoring instances for labeling:
\begin{equation}
\begin{aligned}
    d(x_A, x_B)
    \;=\; \sum_{i \in \{1,2,3,\text{vis}\}} \lambda_i \cdot d_{\cos}\!\big(f_i(x_A), f_i(x_B)\big).
\end{aligned}
\end{equation}
Because the visual embeddings are obtained from frozen models (no additional training), the added computation is negligible relative to detector training. Overall, our diversity-based selection is straightforward to implement, shown in experiments to be effective, and markedly more training-time efficient than conventional ensembles.

\section{Experiments}
\label{sec:results}

\subsection{Datasets, Metrics, and Active Learning Setting}
We evaluate on three benchmarks for monocular 3D detection: KITTI~\cite{kitti} and Waymo monocular~\cite{Waymo,caddn} (both vehicle-mounted), and Rope3D~\cite{rope3d} (traffic scenes with diverse camera perspectives~\cite{groundmix}). Waymo and Rope3D are large-scale datasets. MonoLSS~\cite{monolss} on KITTI \cite{kitti} and MonoCon~\cite{monocon} on Waymo \cite{Waymo} and Rope3D \cite{rope3d} are used as baseline detectors.

\begin{itemize}
    \item \textbf{KITTI~\cite{kitti}.} Following~\cite{chen20153d,monolss}, we split the training set into 3{,}712 images for training, 3{,}769 for validation, and 7{,}518 for testing. We report $AP_{3D|R_{40}}$~\cite{3dr40} (3D Average Precision at 40 recall positions) with the standard Easy/Moderate/Hard difficulty levels, using Moderate as the primary benchmark. For active learning, we start from an initial 10\% labeled training pool and increase the labeled set through iterations to 15\%, 20\%, 30\%, 40\%, 50\%, and 60\%.
    
    \item \textbf{Waymo~\cite{Waymo}.} Following the monocular setting~\cite{caddn}, we use front-camera images with 52{,}386 training and 39{,}848 validation images. We report $AP_{3D}$ and $APH_{3D}$ at IoU thresholds 0.5 and 0.7, and follow MonoLiG~\cite{monolig} in using difficulty level 2. We begin with 10\% labeled images and acquire an additional 5\% per iteration up to a 25\% maximum budget.

    \item \textbf{Rope3D~\cite{rope3d}.} We use the heterologous split (changing camera views), comprising 40{,}333 training and 4{,}676 validation images. Performance is evaluated using $AP_{3D|R_{40}}$~\cite{3dr40} and the Rope Score~\cite{rope3d} at IoU 0.5 and Moderate difficulty. We begin with 20\% labeled images and acquire an additional 5\% per iteration up to a 35\% maximum budget.
\end{itemize}

\begin{table*}[t]
    \begin{minipage}{1.0\textwidth}
    \caption{\textbf{\ac{AL} performance on KITTI \cite{kitti} validation, Waymo \cite{Waymo,caddn} validation and Rope3D \cite{rope3d} validation dataset}. Results are averaged over three rounds, each initialized from the same checkpoint. \textbf{KITTI}: We report \metricRequested$_{60\%}$. 
    \textbf{Waymo}: We report \metricRequested$_{25\%}$.
    \textbf{Rope3D}: We report \metricRequested$_{35\%}$. The appendix provides a description of each method. \textbf{Type$^*$}: U=Uncertainty-based, D=Diversity-based, H=Hybrid.}
    \vspace{-0.6em}
    \centering
    \fontsize{8.8}{11.4}\selectfont
    \begin{tabularx}{\linewidth}{
        @{}c X
        c
        S[table-format=2.2]@{\hspace{0.8em}}
        S[table-format=2.2]@{\hspace{0.8em}}
        S[table-format=2.2]@{\hspace{0.8em}}
        c
        S[table-format=1.2]@{\hspace{0.2em}}
        S[table-format=1.2]@{\hspace{0.2em}}
        S[table-format=1.2]@{\hspace{0.2em}}
        S[table-format=1.2]@{\hspace{0.2em}}
        c
        S[table-format=2.2]@{\hspace{0.6em}}
        S[table-format=2.2]@{\hspace{0.6em}}
        S[table-format=2.2]@{\hspace{0.6em}}
        S[table-format=2.2]@{}
    }
      \toprule

     & & & \multicolumn{3}{c}{\textbf{KITTI \cite{kitti} Car}}
      & \; & \multicolumn{4}{c}{\textbf{Waymo \cite{Waymo} Vehicle}} 
      & \; & \multicolumn{4}{c}{\textbf{Rope3D \cite{rope3d}}} \\
        
     \cmidrule(lr){4-6} \cmidrule(lr){8-11} \cmidrule(lr){13-16}
     &  & & \multicolumn{3}{c}{{$AP_{3D|R_{40}}^{0.7}$}} & & \multicolumn{2}{c}{{IoU=0.5}} & \multicolumn{2}{c}{{IoU=0.7}} & & \multicolumn{2}{c}{{Car}} & \multicolumn{2}{c}{{Big Vehicle}} \\
       
     \cmidrule(lr){4-6} \cmidrule(lr){8-9} \cmidrule{10-11} \cmidrule(lr){13-14} \cmidrule(lr){15-16}
     & Method & Type$^*$ & {Easy} & {Mod.} & {Hard}  & & {AP$_\text{3D}$} & {APH$_\text{3D}$} & {AP$_\text{3D}$} & {APH$_\text{3D}$} & & {AP$_\text{3D}^{0.5}$} & {Rope}& {AP$_\text{3D}^{0.5}$} & {Rope} \\

      \midrule
      
      \multirow{8}{*}{\rotatebox{90}{\textbf{Image-based}}} 
      & Rand & - & 18.01 & 12.97 & 10.77 & & 8.07 & 8.00 & 1.72 & 1.70 &
      & 32.35 &  44.84 & 13.65 & 28.27 \\

      & Conf & U & {\underline{19.72}} & {\underline{14.20}} & {\underline{11.80}} & & 8.03 & 7.96 & 1.74 & 1.72 & & 31.80 & 44.43 & 14.05 & 28.77 \\

      & Ens Depth Var & U & 18.19 & 13.00 & 10.77 & & 7.68 & 7.61 & 1.62 & 1.61 &  & 32.89 & 45.29 & 14.44 & 28.92 \\

      & Augm Depth Var & U & 18.24 & 13.16 & 10.90 & & {-} & {-} & {-} & {-} & & {-} & {-} & {-} & {-} \\

      & Core-Set  \cite{core_set} & D & 18.80 & 13.50 & 11.34 & & 8.10 & 8.03 & 1.76 & 1.75 & & 32.21 & 44.73 & 14.48 & 29.09 \\

      & BADGE \cite{badge} & H & 19.02 & 13.53 & 11.39 & & 8.14 & 8.08 & 1.76 & 1.75 & & 32.83 & 45.22 & 14.05 & 28.61 \\

      & DDFH \cite{ddfh} & H & 17.94 & 12.72 & 10.51 & & 4.50 & 4.46 & 0.97 & 0.96 & & 20.06 & 27.55 & 10.11 & 18.67 \\
      
      & CDAL \cite{cdal} & H & 18.38 & 12.99 & 10.82 & & 7.80 & 7.73 & 1.67 & 1.66 & & 30.96 & 42.43 & 12.18 & 25.97 \\

      \midrule

      \multirow{8}{*}{\rotatebox{90}{\textbf{Instance-based}}} 
      & Rand & - & 18.69 & 13.53 & 11.21 & & 8.19 & 8.12 & 1.85 & 1.84 & & 30.71 & 43.55 & 10.75 & 26.14 \\

      & Conf & U & 12.10 & 8.93 & 7.60 & & 8.06 & 8.00 & 1.83 & 1.82 & & 32.50 & 44.97 & 11.77 & 26.87 \\

      & Ens Depth Var & U & 11.88 & 9.07 & 7.82 & & 8.79 & 8.72 & 2.05 & 2.04 & & 32.43 & 44.90 & 13.56 & 28.32 \\

      & Augm Depth Var & U & 12.25 & 8.96 & 7.64 & & {-} & {-} & {-} & {-} & & {-} & {-} & {-} & {-} \\

      & ComPAS \cite{ComPAS} & U & 17.31 & 12.31 & 10.48 & & 8.63 & 8.56 & 1.96 & 1.94 & & 32.66 & 45.12 & 16.95 & 26.87 \\

      & Core-Set \cite{core_set} Box$_{\text{3D}}$ & D & 18.83 & 13.82 & 11.68 & & 8.84 & 8.77 & 2.02 & 2.01 & & 32.66 & 45.09 & {\underline{16.95}} & {\underline{31.12}} \\

      & BADGE \cite{badge} & H & 17.39 & 12.63 & 10.74 & & {\underline{8.86}} & {\underline{8.79}} & {\underline{2.07}} & {\underline{2.06}} & & {\underline{32.92}} & {\underline{45.32}} & 13.76 & 28.58 \\

      & \ourmethod ~{\scriptsize (Ours)}& D & {\textbf{22.74}} & {\textbf{16.18}} & {\textbf{13.57}} & & {\textbf{9.02}} & {\textbf{8.91}} & {\textbf{2.11}} & {\textbf{2.10}} & & {\textbf{34.27}} & {\textbf{46.41}} & {\textbf{18.45}} &  {\textbf{32.34}} \\
      
      \bottomrule
    \end{tabularx}
    \label{tab:main_results}
    \end{minipage}%
    \vspace{-0.8em}
\end{table*}
\begin{figure*}[t]
  \centering
   \includegraphics[width=1.0\textwidth]{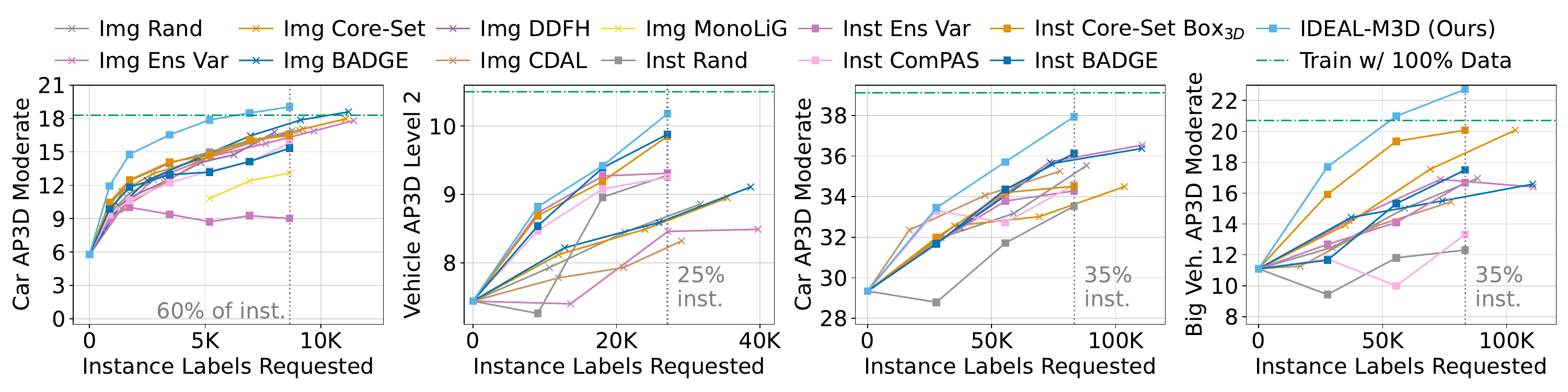}
    \vspace{-0.8cm}
  \caption{\textbf{\ac{AL} training curves.} \textbf{Plot 1}: We report the KITTI validation ~\cite{kitti} performance 
  for cars $AP^{0.7}_{3D|R_{40}}$ Moderate. \textbf{Plot 2}: We report the Waymo validation ~\cite{Waymo} performance $AP^{0.5}$ for vehicles. 
  \textbf{Plots 3-4}: We report the Rope3D validation performance $AP_{3D|R_{40}}^{0.5}$ for Cars and Big Vehicles. All results show mean performance across three rounds with identical initialization.}
  \label{fig:main_results_curves}
  \vspace{-0.5cm}
\end{figure*}

\subsection{NAURC: A budget-fair evaluation metric}
The goal of \ac{AL} is to maximize accuracy under a limited labeling budget. For comparability with prior work, we report our main results using the two standard strategies: training curves over requested instances~\cite{ComPAS,plug_play,mi-aod,monolig,core_set,disc_al,comp_al} and performance at fixed labeled-data percentages \cite{efficient_al,ComPAS,stone}. These approaches, however, have notable drawbacks: training curves can be ambiguous because leadership changes across the budget, and fixed-percentage snapshots provide only a single number that depends on the chosen percentage. They also do not resolve the budget-unit mismatch between image-based methods (spending in images) and instance-based methods (spending in instances).

To address these issues, we propose a new metric, \textbf{\acf{NAURC}}. \ac{NAURC} is a single-scalar metric that integrates performance over the requested-instance budget, normalizes by a target budget, and enables direct comparison between image- and instance-based methods. It enables a common instance-based accounting for all methods: If an image-based method overshoots the target budget, we interpolate back to the budget; if it undershoots, we keep the last observed performance. We refer to the \supp for a formal definition and more detailed motivation, derivation, and visualizations

.

\subsection{Comparison with \ac{AL} methods}
\label{sec:main_results}



In \cref{tab:main_results} and \cref{fig:main_results_curves}, we report average accuracy over three runs initialized from the same checkpoint, comparing \ac{IDEAL-M3D} with other \ac{AL} methods. Interestingly, methods based on uncertainty significantly drop their KITTI \cite{kitti} and Rope3D \cite{rope3d} performance when they move from image-based to instance-based approaches. Intuitively, this is a result of their bias toward distant objects (\cref{sec:div_vs_unc_methods}), which can be mitigated by annotating all the objects in the image. Remarkably, with only 60\% of instances, we surpass the 100\% KITTI baseline, and on Rope3D we reach 97\% (car) and 107\% (big vehicle) of fully supervised AP performance using just 35\% of labels. Results for the pedestrian and cyclist class are shown in the supplementary material.  

\begin{figure*}
  \centering
    \includegraphics[width=\textwidth]{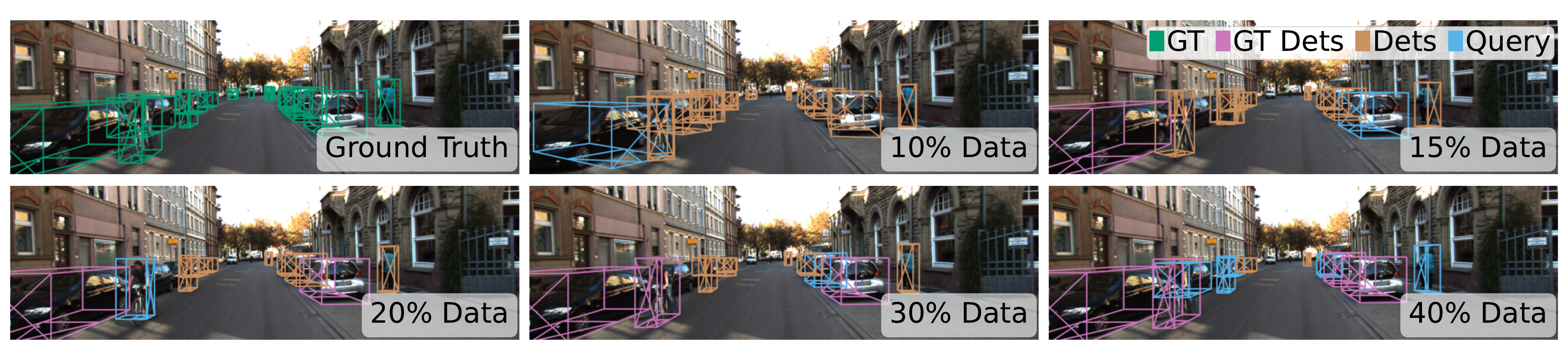}\vspace{-1em}
    \includegraphics[width=\textwidth]{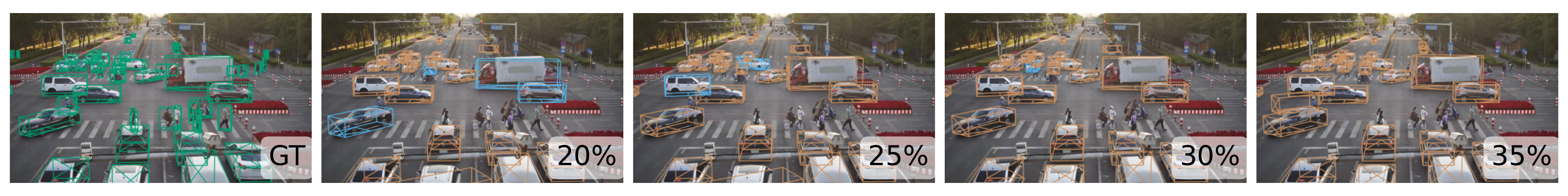}\vspace{-1em}
  \caption{
\textbf{Qualitative results of \ac{IDEAL-M3D} on the KITTI ~\cite{kitti} (first/second row), 
and Rope3D ~\cite{rope3d} (third row) datasets}. The results demonstrate prediction evolution and label selection strategy across time steps. Color coding: green boxes represent ground truth annotations, pink boxes indicate predictions on previously labeled objects, cyan boxes highlight predictions selected for the next labeling round, and orange boxes show predictions that remain unlabeled (best viewed in color with zoom).
}
\vspace{-1.3em}
  \label{fig:05_main_qualit_results}
\end{figure*}

Waymo \cite{Waymo} shows the opposite trend: all instance-based methods improve. We hypothesize that this stems from strong redundancy in the dataset, as Waymo provides video frames at 300 ms intervals \cite{caddn}, whereas KITTI and Rope3D are image-based. On Waymo, we match 97\% (AP$^{0.5}$) and 95\% (AP$^{0.7}$) supervised performance with 75\% fewer labels. These quantitative results are supported by qualitative evidence in \cref{fig:05_main_qualit_results}. We observe that \ac{IDEAL-M3D} tends to prioritize nearby objects at the beginning of \ac{AL} and gradually moves to more distant ones. Additionally, the detection quality generally shows consistent improvement over time.


\subsection{Comparison with fully supervised methods}
\label{sec:comparison_fully_supervised}
\begin{table} 
\caption{Comparison with \ac{SOTA} monocular methods on the KITTI \cite{kitti} test set for the car category. \textbf{SSL} denotes methods that additionally require unlabeled data to perform semi-supervised learning. \textbf{AL} denotes methods that use active learning.}
\vspace{-0.6em}
\scriptsize \centering \resizebox{\columnwidth}{!}{ 
\begin{tabular}{@{}l@{\hspace{2pt}}|c@{\hspace{2pt}}r@{\hspace{2pt}}|c@{\hspace{2pt}}@{\hspace{2pt}}r@{\hspace{2pt}}r@{\hspace{2pt}}r@{}} \toprule

      \multirow{2}{*}{Method}& SSL & Train & \multicolumn{3}{c}{Car Test AP$_{3D|R40}^{70}$}\\
      & /AL & {data} & {Easy} & {Mod} & {Hard} \\ \midrule

    Mix-Teaching,
    \cite{mix-teaching}, 
    TCSVT 23 & \cmark/\xmark & 100\% & \underline{26.89} & 18.54 & 15.79 \\
    DPL$_{\text{FLEX}}$ \cite{decoupled_ssl}, 
    CVPR 24 & \cmark/\xmark & 100\% & 24.19 & 16.67 & 13.83 \\
    MonoLiG \cite{monolig}, WACV 24 & \cmark/\cmark & 100\% & 24.90 & 18.86 & \underline{16.79} \\
    MonoCD \cite{mono_cd}, CVPR 24 & \xmark/\xmark & 100\% & 25.53 & 16.59 & 14.53 \\
    MonoMAE \cite{monomae}, NIPS 24 & \xmark/\xmark & 100\% & 25.60 & 18.84 & 16.78 \\
    MonoDGP \cite{monodgp}, 
    CVPR 25 & \xmark/\xmark & 100\% & 26.35 & 18.72 & 15.97 \\
    GATE3D \cite{gate3d}, CVPRW 25 & \xmark /\xmark & 100\% & 26.07 & 18.85 & 16.76 \\
    \midrule
    MonoLSS, 3DV 24 \scriptsize{(Baseline)} & \xmark/\xmark & 100\% & 26.11 & \textbf{19.15} & \textbf{16.94} \\
        
    \ourmethod 60\% \scriptsize{(Ours)} & \xmark/\cmark & \textbf{60\%} & \textbf{27.06} & \underline{18.87} & 16.73 \\
    
    \bottomrule
    \end{tabular}
    }
    \vspace{-1.4em}
\label{tab:kitti_test}
\end{table}

To assess generalizability, we run active learning up to 60\% of the KITTI trainval instances and evaluate on the KITTI test set~\cite{kitti}. Test labels are not public, so evaluation requires submission to the official server. Competing fully supervised baselines are trained on 100\% of trainval.

In \cref{tab:kitti_test}, we compare \ourmethod with recent fully supervised and semi-supervised approaches. Relative to our fully trained baseline detector MonoLSS~\cite{monolss}, \ourmethod achieves slightly lower AP on Moderate ($-0.28$) and Hard ($-0.21$). Yet, it attains higher AP on Easy ($+0.95$) while using 40\% fewer labels. \ourmethod is competitive with state-of-the-art under substantially reduced annotation. Several entries are semi-supervised: Mix-Teaching~\cite{mix-teaching}, DPL$_{\text{FLEX}}$~\cite{decoupled_ssl}, and MonoLiG~\cite{monolig} leverage additional unlabeled data during training. MonoLiG \cite{monolig} further uses LiDAR during training. In contrast, \ourmethod uses RGB images only.

\subsection{Ablation Study}
\label{sec:ablation}

We perform our main ablation study in \cref{tab:main_ablation} using \ac{NAURC}$_{60\%}$ AP Mod. as the primary metric. Surprisingly, moving from image-based Core-Set~\cite{core_set} to a naïve instance-based Core-Set lowers performance from 13.50 to 13.16 AP Mod. We hypothesize two causes. First, monocular 3D detection is dominated by 3D box estimation (especially depth), while penultimate classification features over-emphasize semantics and are invariant to geometry. Second, image-based selection may benefit from an averaging effect: labeling one image annotates all objects, so even suboptimal choices often include informative instances. Aligning the selection space with the detection task, i.e. replacing classification features with pre-head detection features (\coreSetBox), more than recovers this loss.


Building on this representation, the diverse ensemble (\cref{sec:diverse_ensembles}) contributes +0.72 AP Mod. Adding task-agnostic visual embeddings (\cref{sec:task_agnostic_features}) yields a larger +1.27 AP Mod. This shows that appearance cues complement task-conditioned geometry. The visual pathway adds negligible overhead because embeddings are computed with frozen encoders (\cref{tab:exp_training_time_comp}). This preserves our favorable runtime profile. Using both sources together, detector-specific geometry and detector-agnostic appearance, adds a further +1.64 and +1.09 AP Mod. This aligns with our mutual-information view of non-redundant signals. Notice, that in an image-based selection setting, our approach still exceeds the classical Core-Set baseline by +1.33 AP Mod.


\label{sec:results_ablation}
\begin{table}
    \caption{\textbf{Main ablation study on the KITTI \cite{kitti} validation set (Cars, $AP_{3D|R_{40}}$ , IoU=0.7)}. \textbf{Inst}: Instance-based \ac{AL}. \textbf{Box$_\text{3D}$}: Usage of backbone instead of classification features. \textbf{DE}: Diverse Ensemble. \textbf{VD}: Visual diversity. \textbf{Easy/Moderate/Hard}: We report \ac{NAURC}$_{60\%}$. \textbf{Final AP}: Moderate AP after training on 60\% of the data. For image-based methods we report the interpolated result.}
    \vspace{-0.5em}
    \centering
    \small
    \begin{tabularx}{\linewidth}{
        cccc
        r@{\hspace{0.3em}}
        S[table-format=2.2]@{\hspace{0.2em}}
        S[table-format=2.2]@{\hspace{0.2em}}
        S[table-format=2.2]@{\hspace{0.2em}}
        S[table-format=2.2]@{\hspace{0.2em}}
        S[table-format=2.2]
        }
      \toprule
      Inst & Box$_\text{3D}$ & DE & VD & {Easy} & {Moderate} & {Hard} & {Final AP} \\ \midrule

      & & & & 18.80 & 13.50 & 11.40 & 16.75 \\
      \cmark & & & & 18.28  & 13.16 & 10.94 & 16.37 \\ 
      \cmark & \cmark & & & 18.83 & 13.82  & 11.68  & 16.45 \\ 
      \cmark & \cmark & \cmark & & 20.55 & 14.54  & 12.28  & 18.30 \\ 
      \cmark & \cmark & & \cmark & 20.79 & 15.09  & 12.54 & 18.51 \\
      & \cmark & \cmark & \cmark & 20.77 & 14.83  & 12.34  & 17.96 \\ 
      \cmark & & \cmark & \cmark & 20.21 & 14.96 & 12.53 & 17.82 \\
      \cmark & \cmark & \cmark & \cmark & \textbf{22.74} & \textbf{16.18} & \textbf{13.57} & {\textbf{19.04}} \\

      \bottomrule 
      \label{tab:main_ablation}
      \end{tabularx}
      \vspace{-2.4em}
\end{table}

\subsection{Training-time efficiency}
\begin{table}[h]
  \centering
  \vspace{0.4em}
  \caption{Runtime ablation study on the KITTI \cite{kitti} validation set.}
  \vspace{-0.4em}
  \resizebox{\columnwidth}{!}{%
    \begin{tabular}{ll}
      \textbf{Method} & \textbf{Total training time} \\ \midrule
      Inst. \coreSetBox            & \cbarm{teal!65}{2.664em}{0.7em}{18.2h} \\
      Ours w/o ensemble            & \cbarm{teal!65}{2.736em}{0.7em}{18.7h} \\
      Ours w/o visual diversity    & \cbarm{teal!65}{4.8744em}{0.7em}{33.3h} \\
      Ours w/o SAMv2               & \cbarm{teal!65}{4.888em}{0.7em}{33.4h} \\
      \ourmethod \scriptsize{(Ours)}           & \cbarm{teal!95}{4.932em}{0.7em}{33.7h} \\
      Ours w/o diverse backbones   & \cbarm{teal!65}{5.412em}{0.7em}{37.0h} \\
      Ours w/o data sampling       & \cbarm{teal!65}{5.988em}{0.7em}{40.9h} \\
      Ours w/ full epochs         & \cbarm{teal!65}{6.048em}{0.7em}{41.3h} \\
      \midrule
      Inst Ens Depth Var                  & \cbarm{teal!65}{7.92em}{0.7em}{54.7h} \\

    \end{tabular}
  }
  \vspace{-1.1em}
\label{tab:exp_training_time_comp}
\end{table}\label{sec:training_time_efficiency}
In \cref{tab:exp_training_time_comp}, we report the training-time ablation on the KITTI validation set~\cite{kitti}. The \coreSetBox baseline trains for 18.2h. IDEAL-M3D requires 33.7h in total. Using an ensemble of three models therefore increases training time by only 85\% relative to \coreSetBox. In contrast, a vanilla ensemble like \textit{Inst Ens Depth Var} would naively incur a 200\% increase (approximately $3\times$ the baseline). Among our efficiency measures, shortening the training schedule (fewer epochs) is the most effective. Time-dependent bagging (subsampling more aggressively later) provides the next-largest savings. Smaller, heterogeneous backbones further reduce cost. Together, these measures offset most of the multi-model overhead. Finally, adding task-agnostic visual diversity increases total time by only 1\% because the visual embeddings are computed with frozen models and require no additional detector training.
\section{Conclusion}
We introduced \ac{IDEAL-M3D}, the first instance-based active learning framework for monocular 3D object detection. We show that diversity-driven ensembles are highly effective for instance selection when representational diversity is explicitly maximized. Our design relies on simple mechanisms that increase diversity without heavy training overhead. We also integrate appearance coverage via task-agnostic visual embeddings at negligible cost. Overall, our approach improves label efficiency while keeping computation modest.

Extensive experiments support these claims across three datasets: KITTI, Waymo, and Rope3D, including results on the private KITTI test set. On KITTI, we match fully supervised accuracy using only 60\% of the labeled data. The pipeline is simple to implement and robust in practice.

A limitation is that our approach does not explicitly address extreme class imbalance, where ultra-rare categories may be undersampled. Future work will incorporate multi-modal cues (e.g., LiDAR-derived multi-view images, video) to further diversify signals, and explore semi-supervised learning where feature similarity can guide high-quality pseudo-label selection. \newline

{
\footnotesize
\inparagraph{Acknowledgments}
This work is a result of the joint research project STADT:up. The project is supported by the German Federal Ministry for Economic Affairs and Climate Action (BMWK), based on a decision of the German Bundestag. The author is solely responsible for the content of this publication. This work was also supported by the ERC Advanced Grant SIMULACRON, the Georg Nemetschek Institute project AI4TWINNING and the DFG project 4D-YouTube CR 250/26-1.}

\clearpage

{\small
\bibliographystyle{ieee_fullname}
\bibliography{11_references}
}

\newpage
\begin{center}
\begin{minipage}{0.48\textwidth}  
\setlength{\parskip}{0pt}         
\makeatletter
\def\@dotsep{10000}    
\def\@pnumwidth{0pt}   
\makeatother
{
\tableofcontents
}
\end{minipage}
\end{center}

\section{IDEAL-M3D: Further Details}

\subsection{Problem Statement}

\subsubsection{Monocular 3D Object Detection (M3D).}
\acf{M3D} predicts categories and 3D bounding boxes \(\mathcal{B}_i\) for objects in an RGB image \(I\) with intrinsics \(K \in \mathbb{R}^{3 \times 4}\). Each \(\mathcal{B}_i\) is parameterized by position \((x_i, y_i, z_i) \in \mathbb{R}^3\), dimensions \((w_i, h_i, l_i) \in \mathbb{R}^3\), orientation \(R_i \in SO(3)\), and class \(c_i \in \mathbb{N}\). Given a dataset \(\{(I_j, K_j, \mathcal{B}(I_j))\}_{j=1}^M\), with \(\mathcal{B}(I_j)\) as ground-truth boxes for image \(I_j\), the goal is to train a model capable of predicting \(\mathcal{B}(I)\) for any given image \(I\). Starting solely from a 2D RGB image poses a significant challenge due to depth ambiguity. 

\subsubsection{Active Learning.}
\acf{AL} starts with a small labeled dataset \(\mathcal{D}_L^0 = \{(I_i, K_i, \mathcal{B}(I_i))\}\) and a large unlabeled dataset \(\mathcal{D}_U = \{(I_j, K_j)\}\). In each round \(r\), we select \(\mathcal{D}_r^* = \{(I_j, K_j, \mathcal{S}_{j,r}^*)\}_{j \in \mathcal{J}_r^*}\), where \(\mathcal{J}_r^*\) is the set of selected images and \(\mathcal{S}_{j,r}^* \subseteq \hat{\mathcal{B}}(I_j)\) the subset of bounding boxes chosen for labeling. The oracle \(\Omega: (I, \hat{\mathcal{B}}) \mapsto \mathcal{B} \cup \{\text{null}\}\), which in practice corresponds to a human annotator, refines each selected \(\hat{\mathcal{B}}_{j,k} \in \mathcal{S}_{j,r}^*\), returning \(\mathcal{B}_{j,k}\) or, in case it is not labeled, \(\text{null}\). The labeled dataset is updated as \(\mathcal{D}_L^r = \mathcal{D}_L^0\;\cup\;\bigcup_{r'=1}^r \{(I_j, K_j, \{\mathcal{B}_{j,k}\}_{k \in \mathcal{S}_{j,r'}^*}) \mid \hat{\mathcal{B}}_{j,k} \neq \text{null}\}\), and the model is fine-tuned on \(\mathcal{D}_L^r\). This repeats until the total labeled bounding boxes reach the budget \(\mathcal{T} = \sum_{r=1}^R \sum_{j \in \mathcal{J}_r^*} |\mathcal{S}_{j,r}^*|\). In summary, \ac{AL} pipelines can be seen as iterative cycles that repeat two steps: data selection for labeling, and training. 

\subsection{Loss Functions}
We detail the loss computations for our baseline methods and omit the loss weights for clarity.

\textbf{MonoLSS \cite{monolss}:}
\begin{equation}
    \begin{split}
        L = & L_{cls} + L_{c,o} + L_{h,w} + L_{S_{3d}} + \\
        & L_{\theta} + L_{depth} \cdot \text{Sample } S
    \end{split}
\end{equation}

\textbf{MonoCon \cite{monocon}:}
\begin{equation}
    \begin{split}
        L = & L_{cls} + L_{c,o} + L_{h,w} + L_{depth} + L_{S_{3d}} + \\
        & L_{\theta} + L_{kp,h} + L_{kp,o} + L_{kp,co}
    \end{split}
\end{equation}

The individual losses can be categorized into image- and object-related losses. 

\subsubsection{Image-level Losses}
The following losses are image-specific. Therefore, we apply the masking strategy as described in \cf \cref{sec:04_method_instance_based} on both of these losses:
\begin{itemize}
    \item $L_{cls}$: Gaussian kernel weighted focal loss for classification, following CenterNet ~\cite{centernet}.
    \item $L_{kp,h}$: (MonoCon only) Gaussian weighted focal loss for projected 3D keypoints as an auxiliary task. 
\end{itemize}

\subsubsection{Object-level Losses}
These losses are specific to objects and do not require specialized masking:
\begin{itemize}
    \item $L_{c,o}$: L1 Loss for offset from most confident foreground bin to precise projected 3D center
    \item $L_{h,w}$: L1 loss for 2D height and width
    \item $L_{S_{3d}}$: Dimension-loss. Dimension-aware L1 loss (L1 loss normalized by ground truth) in case of MonoCon and L1 loss in case MonoLSS
    \item $L_{depth}$: Laplacian aleatoric uncertainty loss
    \item $L_{\theta}$: Multi-bin loss following Mousavian et al. ~\cite{multi_bin_loss}
    \item $L_{kp,o}$: L1 loss for keypoint offsets from keypoint heatmap
    \item $L_{kp,co}$: L1 loss for keypoint offsets from projected 3D center
\end{itemize}

\subsection{Obtaining Diverse Features}
\paragraph{Ensemble Features}
When extracting features for \coreSetBox we orientate on our baseline detectors. The idea is simple: We use the features that lead to a bounding box prediction. For MonoLSS ~\cite{monolss} we use the \ac{RoI} features of dimension $d$ x 7 x 7, while for MonoCon ~\cite{monocon} we use the features of size $d$ x 3 x 3. Before applying Core-Set \cite{core_set} selection, the tensors are flattened into a single dimensional vector. 

For our main model we employ the standard DLA-34 ~\cite{dla} backbone ($d=64$). For our auxiliary models we replace the DLA-34 with the backbones of RepVIT M ~\cite{repvit} ($d=56$) and MobileNetv4 M ~\cite{mobilenetv4} ($d=48$). These models are very lightweight in terms of parameters and still offer an acceptable 2D and 3D detection performance, while being easily interchangeable with minimal code modifications (\cf \cref{fig:backbone_ablation}).

\paragraph{Visual Features}
The extraction of visual features follows a two step process. First we mask the image, then we encode it using an off-the-shelf image autoencoder. 

To ensure compatibility with Core-Set selection, all object features must share the same dimensionality, yet the pixel space of instances differs. For example, a car closer to the camera has a larger pixel height than a car further away. While resizing objects to a fixed size is a straightforward solution, we adopt a more effective strategy (\cf \cref{tab:ablation_resizing}). Specifically, we crop each object to a fixed height and width of 320$\times$320 pixels, centering the crop on the 2D center of the object. If the object lies near the image boundary, we apply padding using the background color. We resize the cropped region to 128x128 pixels before feeding it into the autoencoder. We utilize the autoencoder of Stable Diffusion v2-base ~\cite{ldm} for this purpose.

Our approach ensures that the visual features effectively capture both the object's 2D size and depth through the fixed cropping strategy, which preserves relative scale and encourages depth-diversity. Additionally, the resized crop maintains the object's visual characteristics, enabling the autoencoder to learn a rich, low-dimensional representation of the visual appearance.

For segmentation ~\cite{sam}, we utilize the SAMv2 ViT-B \cite{sam} architecture on Rope3D ~\cite{rope3d} and Waymo \cite{Waymo} and the larger SAMv2 ViT-L \cite{sam} variant on KITTI ~\cite{kitti}.

\begin{table}
    \caption{\textbf{Ablation study on the KITTI ~\cite{kitti} validation set showing the effect of resizing}. Results are reported using \ac{NAURC}$_{60\%}$ $AP_{3D|R_{40}}$ for cars (IoU 0.7) and pedestrians/cyclists (IoU 0.5). \textbf{Final AP}: Moderate AP after training on 60\% of the data.}
    \centering
    \small
    \begin{tabularx}{\linewidth}{
        @{}X 
        r@{\hspace{0.5em}}
        S[table-format=2.2]@{\hspace{0.2em}}
        S[table-format=2.2]@{\hspace{0.2em}}
        S[table-format=2.2]@{\hspace{0.2em}}
        S[table-format=2.1]
        }
      \toprule
      Method & {Easy} & {Moderate} & {Hard} & {Final AP} \\ \midrule
      
      Ours w/ object resizing & 21.51  & 15.17  & 12.55 & 18.59 \\       
      \ourmethod (\scriptsize{Ours}) & \textbf{22.74} & \textbf{16.18} & \textbf{13.57} & {\textbf{19.04}} \\ 

      \bottomrule 
      \label{tab:ablation_resizing}
      \end{tabularx}
\end{table}

\section{Experiments}
\subsection{NAURC Evaluation Metric}
\label{sec:metric_naurc_detail}

A central challenge in \ac{AL} evaluation is to obtain a scalar, budget-aware summary that is comparable across selection paradigms. Previous work typically use two approaches: (i) training curves that plot performance against the number of labeled instances on the x-axis~\cite{ComPAS,plug_play,mi-aod,monolig,core_set,disc_al,comp_al}; while informative, rankings often swap across budgets/time, complicating objective comparison; and (ii) fixed-budget snapshots that report performance at selected percentages of labeled data~\cite{efficient_al,ComPAS,stone}; these yield a single value but depend on the chosen percentage and ignore the rest of the trajectory. 

We introduce the \acf{NAURC} to provide a fair, single-scalar comparison across image- and instance-based methods. NAURC adopts a common instance-based accounting for all approaches: for image-based selection, each selected image contributes the number of labelable instances it contains (the budget accumulates the total instances from the chosen images); for instance-based selection, the budget counts all requested instances, including requests that are later deemed nonlabelable, thereby reflecting annotator verification effort. NAURC normalizes by a target budget and handles budget mismatch robustly: if a method overshoots the budget, we linearly interpolate back to it; if it undershoots, we keep the last observed performance (no extrapolation).

\begin{figure}[H]
  \centering
  \includegraphics[width=0.48\textwidth]{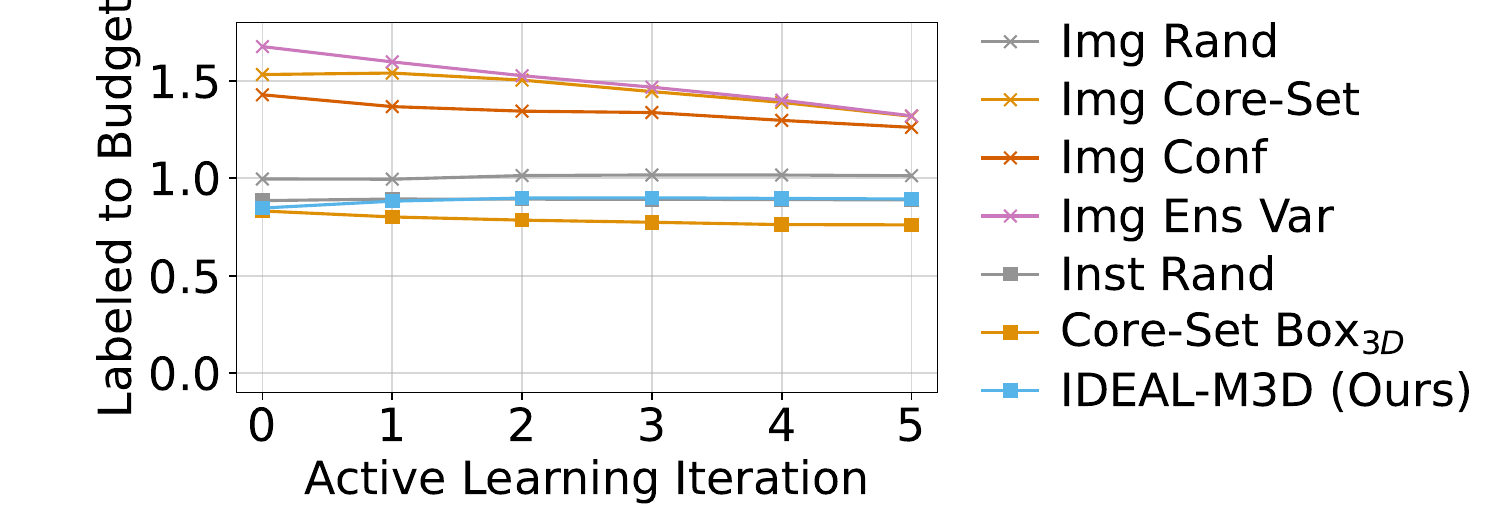}
  \caption{Ratio of labeled instances vs.\ instance-based budget on KITTI~\cite{kitti}. Most image-based methods request images with more than the average number of objects.}
  \label{fig:labeled_ratios}
\end{figure}

\paragraph{Empirical motivation for instance-based accounting.}
We analyze the relationship between actually labeled instances and the allocated budget in \cref{fig:labeled_ratios}. For image-based methods, we quantify the budget at each \ac{AL} iteration in terms of actually labeled boxes, since these methods request whole images. Most image-based strategies, with the exception of \textit{Img Random}, tend to select images containing above-average numbers of instances, revealing that an image-count budget unfairly mixes annotation efforts because images vary widely in object count. For instance-based methods, the labeled ratio captures the proportion of true positive requests. Because false positive requests still require annotator verification, they count against the budget. This slightly penalizes instance-based methods compared to image-based methods. However, it also considers the annotators effort to verify that a requested object is a false positive.

\begin{figure}[H]
  \centering
  \includegraphics[width=0.465\textwidth]{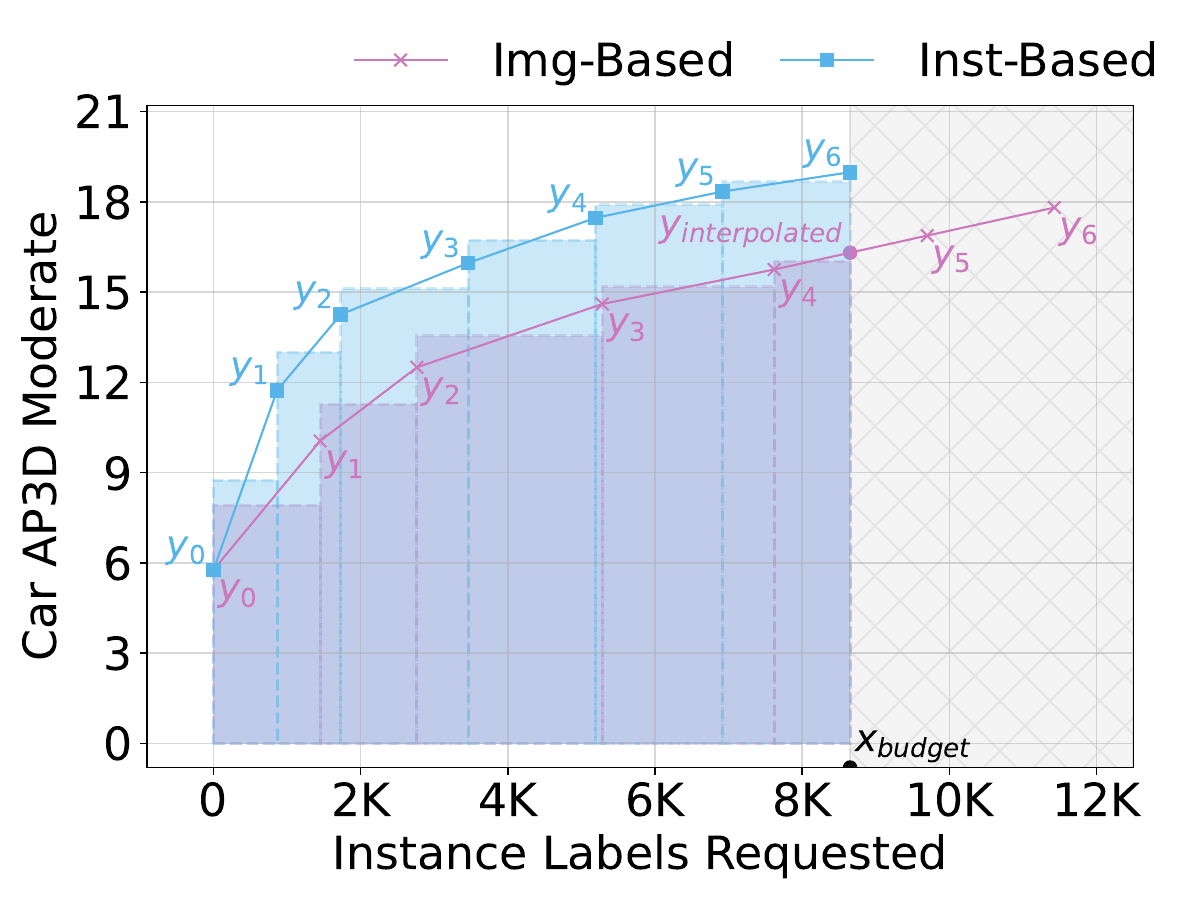}
  \caption{Visualization of the \acl{NAURC} computation. For methods exceeding the label budget $x_{\text{budget}}$, the final performance metric is interpolated.} 
  \label{fig:naurc_metric}
\end{figure}

Formally, we compute the metric as the \ac{AURC} up until the total requested label budget $ x_{\text{budget}} $ normalized by $ x_{\text{budget}} $ (refer to \cref{fig:naurc_metric}). The \ac{NAURC} is defined as:
\begin{equation}
    \text{NAURC}_{x_{\text{budget}}} = \frac{1}{x_{\text{budget}}} \left( \text{AURC}_{\text{final}} + \sum_{i=0}^{k} \text{AURC}_i \right),
\end{equation}
where $ k $ denotes the last \ac{AL} iteration before reaching the requested instance label budget $ x_{\text{budget}} $:
\begin{equation}
    k = \max \left\{ i \mid x_{i+1} \leq x_{\text{budget}} \right\}.
\end{equation}
We compute the metric across iterations using the trapezoidal rule to calculate the \ac{AURC} between consecutive data points.
\begin{equation}
    \text{AURC}_i = \frac{(y_{i+1} + y_i)}{2} \cdot (x_{i+1} - x_i),
\end{equation}
where $ y_i $ and $ y_{i+1} $ are the metric values at the $ i $-th and $ (i+1) $-th \ac{AL} iteration and $ x_i $ and $ x_{i+1} $ are their respective requested instance labels.
The \ac{AURC} of the final interval is computed as:
\begin{equation}
    \text{AURC}_\text{final} = \frac{(y_{\text{interpolated}} + y_k)}{2} \cdot (x_{\text{budget}} - x_k),
\end{equation}
where $ y_{\text{interpolated}} $ is the metric value at the budget point. To handle both cases where methods exceed or fall short of the target budget, we define:
\begin{equation}
    y_{\text{interpolated}} = \begin{cases}
        y_{k} + \Delta y \cdot \frac{x_{\text{budget}} - x_{k}}{x_{k+1} - x_{k}} & \text{if } x_{k+1} > x_{\text{budget}} \\
        y_{k} & \text{otherwise}
    \end{cases}
\end{equation}
where $\Delta y = y_{k+1} - y_{k}$. 

Compared to prior AUC-style metrics~\cite{comp_al,ijcai2021p634,multiple_criteria_al}, \ac{NAURC} enables fair cross-paradigm comparison across image- and instance-based methods: If a method overshoots the budget, \ac{NAURC} linearly interpolates the terminal performance at $x_{budget}$; if it undershoots, it holds the last observed value. This removes extrapolation and prevents artificial inflation. 

\subsection{Implementational Details}

\label{sec:implementational_details}

\paragraph{Active Learning Parameters.}

The distance weights (\cf \cref{eq:distance}) are determined as $\lambda_1 = \lambda_2 = \lambda_3 = \frac{1}{6}$, with $\lambda_{vis}$ set to $\frac{1}{2}$ to balance the contribution of visibility metrics. The Loss weight multiplier parameter is set to $\delta = 0.2$ 
For adaptive sampling, we configure the time-dependent decay parameter $\alpha$ to 3.0 for KITTI~\cite{kitti} and 30.0 for Rope3D~\cite{rope3d} and Waymo \cite{Waymo} to account for the different datase size.
On KITTI~\cite{kitti}, labeling proposals exclude objects with heights below 25 pixels, as ground truth instances are constrained to a minimum height of 25 pixels, which eliminates unsuitable candidates during this process.
All experiments, including run time evaluations, are conducted on a single NVIDIA A40 GPU with 32GB RAM. For instance-based \ac{AL}, we mitigate redundant labeling by skipping requests for objects that fall within 95\% of the radius $(r_x, r_y)$ from a previous request and share the same predicted class. This prevents repeated annotations or requests regarding the same false-positive predictions.
For masking and visual feature extraction, we leverage the 2D bounding box predictions of the main model. To accelerate computation, we apply \ac{PCA} to the extracted features, compressing the dimensions while retaining at least 99\% of the variance.

\paragraph{MonoLSS~\cite{monolss} Configuration.}

The training setup utilizes a batch size of 16 and optimizes the model using the Adam~\cite{adam} optimizer with a weight decay of 1e-5. Starting from a pre-trained checkpoint, each \ac{AL} iteration spans 150 epochs for the main model.
During training, we initialize the learning rate at 1e-3 and decay it by a factor of 0.1 after 60\% and 80\% of the epochs. Additionally, the first cycle incorporates a 5-epoch cosine warmup to stabilize gradient updates.
After the initial phase of training, the LSS module gets activated at the 50th epoch. To enhance model robustness, we apply comprehensive data augmentation techniques, including random horizontal flipping, shifting (W: $\pm$256 pixels, H: $\pm$77 pixels), scaling (0.6-1.4), and MixUp3D using fully labeled images.

\paragraph{MonoCon~\cite{monocon} configuration.}
For MonoCon, we adopt a batch size of 24 and train the model using the AdamW~\cite{adamw} optimizer with a weight decay of 1e-5 and a learning rate of 0.0011. The initial training phase encompasses 90 epochs, followed by an additional 30 epochs for every subsequent \ac{AL} cycle.
To maintain stable training, we implement gradient clipping with a norm of 35 and apply cosine learning rate scheduling. For computational efficiency, images are resized to 960$\times$640 pixels, in line with the settings from~\cite{groundmix}.
The augmentation pipeline integrates a rich variety of transformations, including photometric distortion, random shift ($\pm$ 32 pixels), horizontal flipping, and random cropping (900$\times$550 pixels). Furthermore, for Rope3D~\cite{rope3d}, we learn the $SO(3)$ orientation matrix in alignment with the GroundMix~\cite{groundmix} approach. 

\paragraph{Labeling radius.}
Also, to simplify the task for the labeler, we expect that an object lies within a depth-dependent radius \((r_x, r_y)\), letting the user focus on a small area. Such radius is defined as:
\begin{equation}
    r_x = H \cdot \frac{f_x}{\hat{z}}, \quad r_y = H \cdot \frac{f_y}{\hat{z}},
\end{equation}
where \(H\) is a scaling factor, \((f_x, f_y)\) are the camera focal lengths, and \(\hat{z}\) is the predicted depth. This ensures accurate targeting by accounting for the geometric effects of depth, as pixel-space errors decrease with distance. We define the labeling radius via $H=2.0$, which corresponds to approximately 47 pixels for objects at a 30m distance on KITTI~\cite{kitti}. 

\paragraph{Instance matching between ensemble members.}
Using the predictions from the main model as a reference, we associate predictions from auxiliary models by calculating the 2D bounding box IoU. This computation uses a relaxed threshold of 0.5 to accommodate additional object matches. As auxiliary models are trained with fewer resources, we adjust detection thresholds to compensate for their lower capacity. Specifically, thresholds are reduced from 0.2 to 0.1 for MonoCon~\cite{monocon} and from 0.2 to 0.05 for MonoLSS~\cite{monolss}.
For cases of multiple associations, we prioritize the object with the highest confidence score. This strategy significantly enhances instance coverage, particularly for rare and low-confidence objects, while maintaining diversity and minimizing discard rates.

\begin{figure*}[t]
  \centering
  \begin{subfigure}[b]{1.0\textwidth}
    \includegraphics[width=\textwidth]{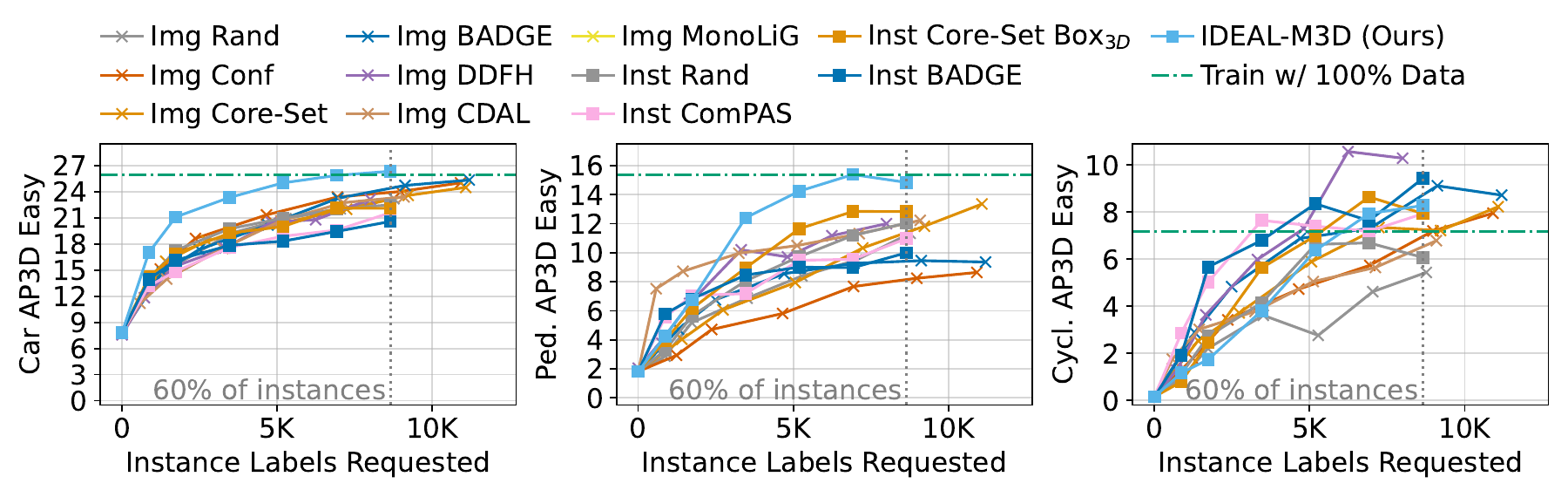}
  \end{subfigure}
  \begin{subfigure}[b]{1.0\textwidth}
    \includegraphics[width=\textwidth]{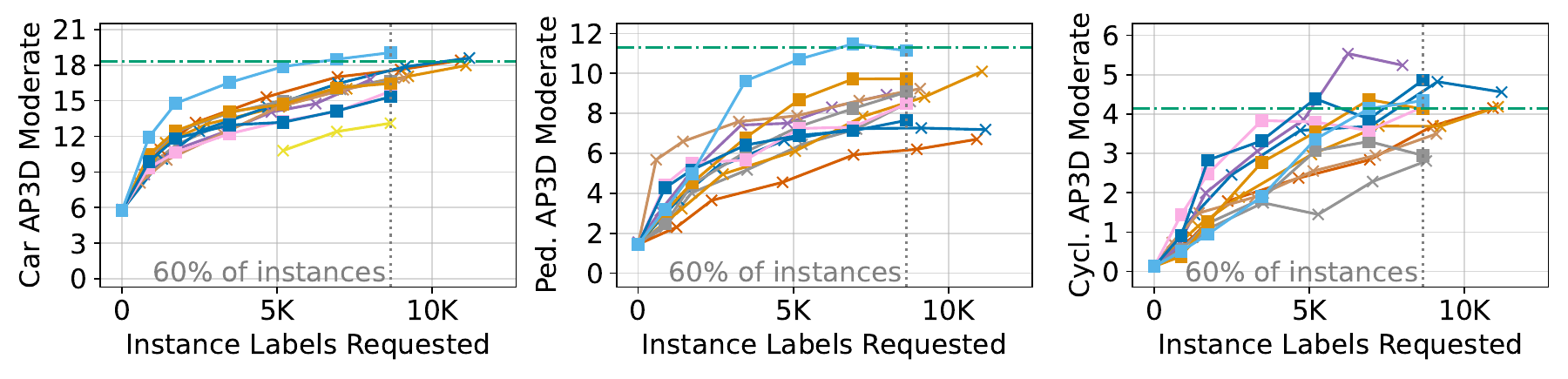}
  \end{subfigure}
  \begin{subfigure}[c]{1.0\textwidth}
    \includegraphics[width=\textwidth]{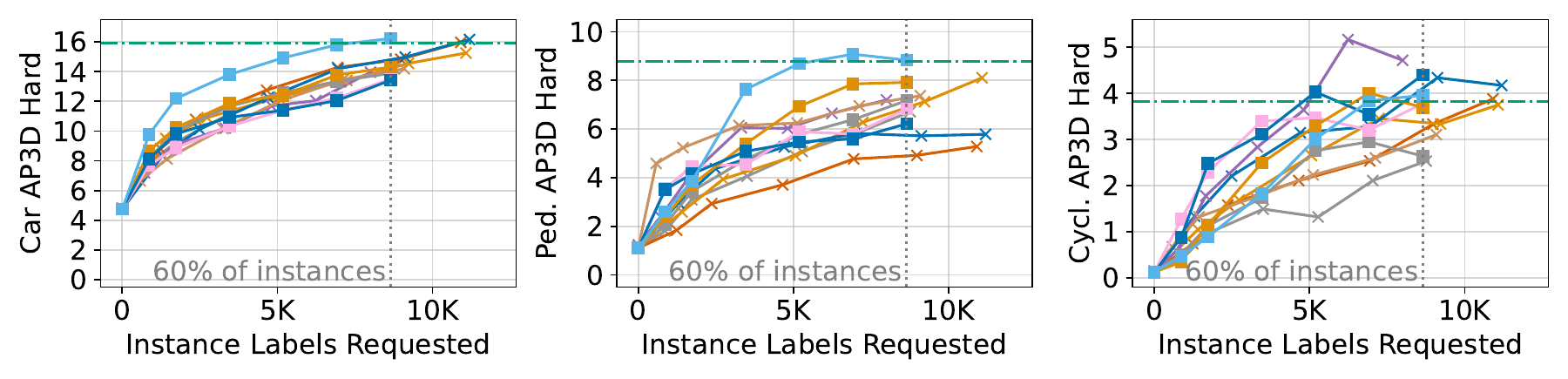}
  \end{subfigure}
  \caption{\textbf{\ac{AL} methods evaluated on the KITTI validation set ~\cite{kitti} for Easy, Moderate and Hard on Car (IoU=0.7), Pedestrian (IoU=0.5) and Cyclist (IoU=0.5)}.}
  \label{fig:12_kitti_result_qualit}
\end{figure*}

\begin{figure*}[t]
  \centering
  \includegraphics[width=1.0\textwidth]{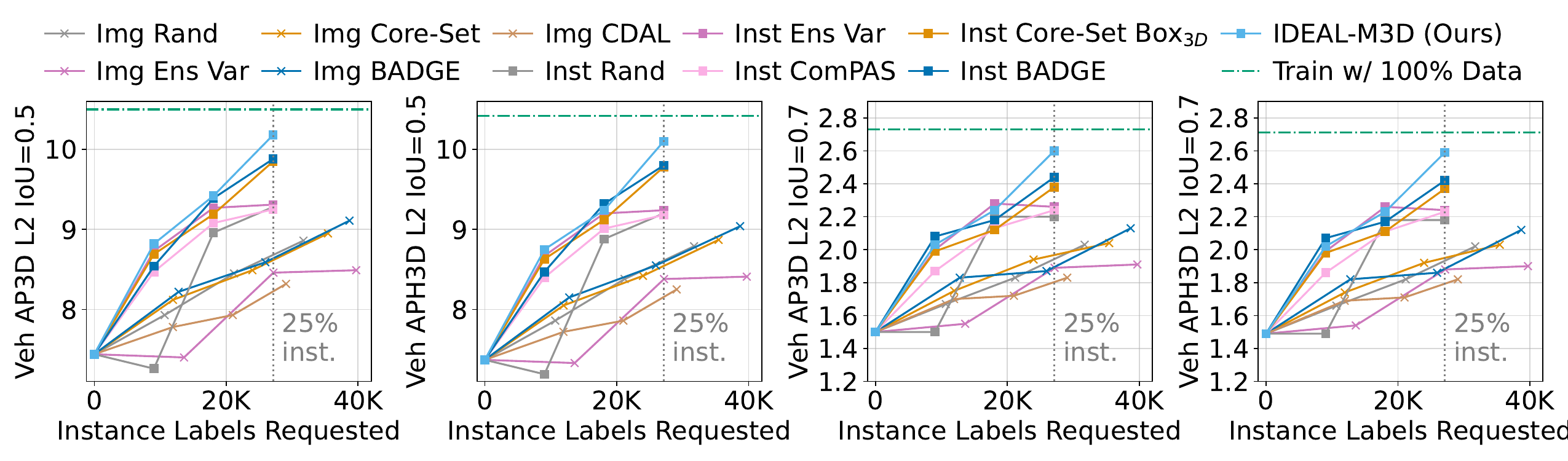}
  \caption{\textbf{\ac{AL} methods evaluated on the Waymo validation set ~\cite{Waymo}}.}
  \label{fig:12_waymo_plots}
\end{figure*}

\begin{figure}[h]
  \centering
  \includegraphics[width=0.49\textwidth]{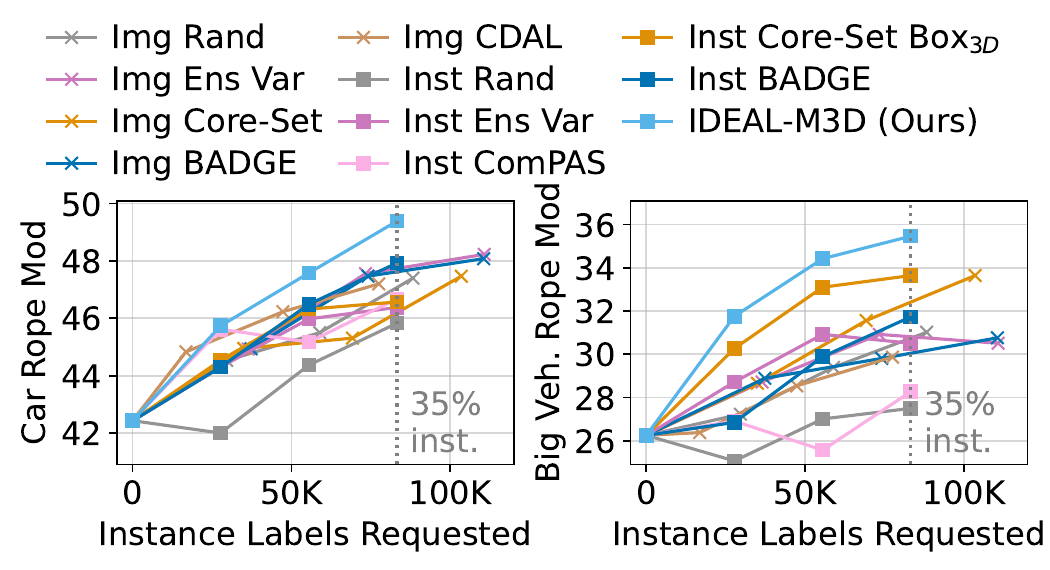}
  \caption{\textbf{\ac{AL} methods evaluated on the Rope3D validation set ~\cite{rope3d}} with Rope Score at IoU 0.5.}
  \label{fig:12_ropescore_plot}
\end{figure}

\subsection{Implementation Details of Baseline Methods}
\label{sec:impl_baseline_mehods}
For completeness, we provide additional implementation details for a subset of our baseline methods. In our evaluation framework, we adapt image-based methods by selecting images based on their highest-scoring contained instance, while instance-based methods directly employ our labeling pipeline (\cref{sec:04_method_instance_based}) with their respective acquisition functions. To manage computational resources effectively, we implement dataset-specific sampling strategies: For datasets containing on average more than 10 objects per image (\eg Rope3D ~\cite{rope3d}), we limit the maximum number of requested images to one-third of the number of instances, though this restriction is not applied to KITTI ~\cite{kitti} due to its limited size. In the following, we provide further details on the individual baseline methods. 

\textbf{Augmentation Depth Variance (Augm Depth Var):} 
This method provides a computationally efficient alternative to ensemble approaches by employing multiple augmented forward passes of a single model. The augmentation pipeline includes blur ($\sigma=0.5$), brightness adjustments (factors: 0.5-1.5, probability: 0.2-0.5), and hue shifts (±0.1). We exclude this method for MonoCon ~\cite{monocon} evaluations due to its built-in color augmentation during training.

\textbf{CDAL ~\cite{cdal}:} 
This approach implements Core-Set sampling by computing class-wise confusions from softmax probabilities of both labeled data heatmap indices and predictions. The sampling strategy employs pairwise and class-wise KL divergence for measuring image similarity.

\textbf{CloseDepth}: Instances with low depth values are preferred for labeling.

\textbf{Confidence (Conf):} 
This uncertainty-based approach combines depth and classification confidence scores from the detector, selecting instances with lower confidence for labeling prioritization. 

\textbf{DDFH} ~\cite{ddfh}: DDFH is the current state-of-the-art method for active learning in LiDAR-based 3D detection. The approach optimizes three key aspects: balancing class distribution, maximizing frame-level heterogeneity, and selecting diverse instances. For instance diversity, it fits a Gaussian mixture model to t-SNE ~\cite{tsne} compressed model features and additional geometric features. We adapt this approach to monocular 3D detection by dropping the point density information from the feature vectors.

\textbf{BADGE ~\cite{badge}:}
BADGE computes gradient embeddings derived from the penultimate classification layer for each data point and selects samples via the k-MEANS++ seeding algorithm ~\cite{kmeanspp_seeding}. We compute gradients based on the focal loss for the predicted heatmap locations on an instance basis.

\textbf{ComPAS ~\cite{ComPAS}:} We adapted the active learning component of ComPAS [citation] to our setting, which selects instances based on their localization and classification disagreement between a chairman and its committee members. This disagreement is measured using multiple data augmentations, including scale, shift, contrast, solarize, saturation, sharpness, and brightness.

\textbf{Dropout-based Methods:}
We investigated dropout-based uncertainty estimation but excluded it from our final evaluation due to significant performance degradation. Specifically, when applying dropout to backbone features, we observed substantial drops in accuracy: While the baseline achieves 18.29 for Car $AP_{3D|R_{40}}^{IoU=0.7}$ Moderate, introducing dropout rates of 1\%, 5\%, and 10\% reduces performance to 17.73, 15.81, and 13.67 on the KITTI ~\cite{kitti} validation set, respectively (baseline: MonoLSS ~\cite{monolss}).

\textbf{Efficient AL ~\cite{efficient_al}:} 
We excluded Efficient AL ~\cite{efficient_al} from our comparison due to ambiguous evaluation metrics (unspecified IoU thresholds for mAP and no differentiation between KITTI~\cite{kitti}'s easy, moderate, and hard instances).

\textbf{Ensemble Depth Variance (Ens Depth Var):} 
Using an ensemble of three models, we associate predictions via a 2D IoU threshold of 0.5. The uncertainty metric is derived from the variance of depth predictions relative to their mean, prioritizing instances exhibiting higher variance.

\textbf{Ensemble Relative Standard Deviation (Ens Rel Std):} Similar to \textit{Ens Depth Var}, but normalizes the depth standard deviation by the mean predicted depth. 

\textbf{FarDepth}: Instances with large depth values are preferred for labeling. Predicted instances need to have a 2D height of at least 25 pixels and a depth smaller 50m.

\textbf{MonoLiG ~\cite{monolig}:} 
MonoLiG ~\cite{monolig} combines \ac{AL} with semi-supervised learning by leveraging an ensemble of five models alongside a LiDAR-based teacher model. The instance selection strategy integrates three uncertainty measures: the aleatoric uncertainty of the teacher model, the disagreement among ensemble members, and the teacher-student prediction inconsistency. These measures are aggregated into a unified scoring mechanism for ranking potential labeling candidates.
Due to unavailability of the official implementation at submission time, we report the results from the paper. To establish a fair comparison, we estimate the number of labeled instances per \ac{AL} iteration based on the dataset's average object count per image.

\subsection{Comparison with \ac{AL} methods}
\begin{table*}[t]
    \begin{minipage}{1.0\textwidth}
    \caption{\textbf{\ac{AL} performance on the KITTI \cite{kitti} validation dataset}. Results are averaged over three rounds, each initialized from the same checkpoint. \textbf{KITTI}: We report \metricRequested$_{60\%}$ $AP_{3D|R_{40}}$ with IoU thresholds of 0.7 (cars) and 0.5 (pedestrians, bicyclists).  \textbf{Type$^*$}: U=Uncertainty-based, D=Diversity-based, H=Hybrid.}
    \centering
    \fontsize{8.8}{11.4}\selectfont
    \begin{tabularx}{\linewidth}{
        @{}c X
        c
        S[table-format=2.2]@{\hspace{0.8em}}
        S[table-format=2.2]@{\hspace{0.8em}}
        S[table-format=2.2]@{\hspace{0.8em}}
        S[table-format=2.2]@{\hspace{0.8em}}
        S[table-format=1.2]@{\hspace{0.8em}}
        S[table-format=1.2]@{\hspace{0.8em}}
        S[table-format=1.2]@{\hspace{0.8em}}
        S[table-format=1.2]@{\hspace{0.8em}}
        S[table-format=1.2]@{\hspace{0.8em}}
        S[table-format=2.2]@{\hspace{0.6em}}
        S[table-format=1.2]@{\hspace{0.6em}}
        S[table-format=1.2]@{}
    }
      \toprule
      
      &  & & \multicolumn{3}{c}{Car} & \multicolumn{3}{c}{Pedestrian} & \multicolumn{3}{c}{Cyclist} & \multicolumn{3}{c}{Average} \\
       
      \cmidrule(lr){4-6} \cmidrule(lr){7-9} \cmidrule(lr){10-12} \cmidrule(lr){13-15} 
     & Method & Type$^*$ & {Easy} & {Mod.} & {Hard} & {Easy} & {Mod.} & {Hard} & {Easy} & {Mod.} & {Hard} & {Easy} & {Mod.} & {Hard} \\

      \midrule
      
      \multirow{8}{*}{\rotatebox{90}{\textbf{Image-based}}} 
      & Rand & - & 18.01 & 12.97 & 10.77 & 6.31 & 4.83 & 3.87 & 2.88 & 1.44 & 1.29 & 9.07 & 6.41 & 5.31\\

      & Conf & U & 19.72 & 14.20 & 11.80 & 5.56 & 4.31 & 3.47 & 4.19 & 2.11 & 1.88 & 9.82 & 6.87 & 5.72 \\

      & Ens Depth Var & U & 18.19 & 13.00 & 10.77 & 4.36 & 3.47 & 2.74 & 4.15 & 2.09 & 1.89 & 8.90 & 6.19 & 5.13 \\

      & Augm Depth Var & U & 18.24 & 13.16 & 10.90 & 4.53 & 3.61 & 2.86 & 3.34 & 1.67 & 1.49 & 8.70 & 6.15 & 5.08 \\

      & Core-Set  \cite{core_set} & D & 18.80 & 13.50 & 11.34 & 7.20 & 5.59 & 4.47 & 4.88 & 2.45 & 2.23 & 10.29 & 7.18 & 6.01 \\

      & BADGE \cite{badge} & H & 19.02 & 13.53 & 11.39 & 7.40 & 5.76 & 4.62 & 5.68 & 2.91 & 2.60 & 10.70 & 7.40 & 6.20 \\

      & DDFH \cite{ddfh} & H & 17.94 & 12.72 & 10.51 & 8.90 & 6.68 & 5.38 & 6.47 & 3.37 & 3.12 & 11.10 & 7.59 & 6.34 \\
      

      & CDAL \cite{cdal} & H & 18.38 & 12.99 & 10.82 & 9.88 & 7.49 & 6.00 & 4.27 & 2.16 & 1.93 & 10.84 & 7.55 & 6.25 \\

      \midrule

      \multirow{8}{*}{\rotatebox{90}{\textbf{Instance-based}}} 
      & Rand & - & 18.69 & 13.53 & 11.21 & 8.03 & 6.05 & 4.72 & 4.49 & 2.10 & 1.91 & 10.40 & 7.23 & 5.95 \\

      & Conf & U & 12.10 & 8.93 & 7.60 & 3.71 & 2.90 & 2.30 & 2.96 & 1.42 & 1.27 & 6.26 & 4.22 & 3.87 \\

      & Ens Depth Var & U & 11.88 & 9.07 & 7.82 & 3.66 & 2.82 & 2.22 & 3.67 & 1.75 & 1.59 & 6.40 & 4.55 & 3.88 \\

      & Augm Depth Var & U & 12.25 & 8.96 & 7.64 & 3.15 & 2.42 & 1.94 & 2.80 & 1.36 & 1.24 & 6.07 & 4.25 & 3.61 \\

      & ComPAS \cite{ComPAS} & U & 17.31 & 12.31 & 10.48 & 8.04 & 6.24 & 5.01 & 6.29 & 3.18 & 2.87 & 10.55 & 7.24 & 6.12 \\

      & Core-Set \cite{core_set} Box$_\text{3D}$ & D & 18.83 & 13.82 & 11.68 & 9.38 & 7.07 & 5.68 & 5.50 & 2.78 & 2.54 & \underline{11.24} & \underline{7.89} & \underline{6.63} \\

      & BADGE \cite{badge} & H & 17.39 & 12.63 & 10.74 & 7.98 & 6.13 & 4.88 & 6.55 & 3.32 & 3.04 & 10.64 & 7.36 & 6.22 \\

      & \ourmethod ~{\scriptsize (Ours)}& D & 22.74 & 16.18 & 13.57 & 11.42 & 8.61 & 6.86 & 4.84 & 2.51 & 2.32 & \textbf{13.00} & \textbf{9.10} & \textbf{7.58} \\
      
      \bottomrule
    \end{tabularx}
    \label{tab:kitti_all_classes}
    \end{minipage}%
    \vspace{-0.8em}
\end{table*}
\begin{table*}[t]
    \begin{minipage}{1.0\textwidth}
    \caption{\textbf{Final \ac{AL} performance on KITTI \cite{kitti} validation, Waymo \cite{Waymo,caddn} valiation and Rope3D \cite{rope3d} validation dataset}. Results are averaged over three rounds, each initialized from the same checkpoint. We report the final accuracy after training on 60\% of data (KITTI) and 25\% of data (Rope3D, Waymo). For image-based methods exceeding the budget we report the interpolated result. \textbf{Type$^*$}: U=Uncertainty-based, D=Diversity-based, H=Hybrid.}
    \centering
    \fontsize{8.8}{11.4}\selectfont
    \begin{tabularx}{\linewidth}{
        @{}c X
        c
        S[table-format=2.2]@{\hspace{0.8em}}
        S[table-format=2.2]@{\hspace{0.8em}}
        S[table-format=2.2]@{\hspace{0.8em}}
        c
        S[table-format=2.2]@{\hspace{0.2em}}
        S[table-format=2.2]@{\hspace{0.2em}}
        S[table-format=1.2]@{\hspace{0.2em}}
        S[table-format=1.2]@{\hspace{0.2em}}
        c
        S[table-format=2.2]@{\hspace{0.6em}}
        S[table-format=2.2]@{\hspace{0.6em}}
        S[table-format=2.2]@{\hspace{0.6em}}
        S[table-format=2.2]@{}
    }
      \toprule

     & & & \multicolumn{3}{c}{\textbf{KITTI \cite{kitti} Car}}
      & \; & \multicolumn{4}{c}{\textbf{Waymo \cite{Waymo} Vehicle}} 
      & \; & \multicolumn{4}{c}{\textbf{Rope3D \cite{rope3d}}} \\
        
      \cmidrule(lr){4-6} \cmidrule(lr){8-11} \cmidrule(lr){13-16}
      &  & & \multicolumn{3}{c}{$AP_{3D|R_{40}}^{0.7}$} & & \multicolumn{2}{c}{IoU=0.5} & \multicolumn{2}{c}{IoU=0.7} & & \multicolumn{2}{c}{Car} & \multicolumn{2}{c}{Big Vehicle} \\
       
      \cmidrule(lr){4-6} \cmidrule(lr){8-9} \cmidrule{10-11} \cmidrule(lr){13-14} \cmidrule(lr){15-16}
     & Method & Type$^*$ & {Easy} & {Mod.} & {Hard}  & & {AP$_\text{3D}$} & {APH$_\text{3D}$} & {AP$_\text{3D}$} & {APH$_\text{3D}$} & & {AP$_\text{3D}^{0.5}$} & {Rope}& {AP$_\text{3D}^{0.5}$} & {Rope} \\

      \midrule
      
      \multirow{8}{*}{\rotatebox{90}{\textbf{Image-based}}} 
      & Rand & - & 22.47 & 16.42 & 13.80 & & 8.68 & 8.20 & 1.74 & 1.73 & & 35.13 & 47.08 & 16.63 & 30.74 \\

      & Conf & U & 23.89 & 17.51 & 14.72 & & 8.12 & 8.05 & 1.66 & 1.65 & & 34.04 & 46.24 & 17.61 & 31.63 \\

      & Ens Depth Var & U & 22.81 & 16.28 & 13.66 & & 8.46 & 8.37 & 8.37 & 1.87 & & 35.93 & 47.74 & 16.76 & 30.81 \\

      & Augm Depth Var & U & 22.20 & 16.15 & 13.72 & & {-} & {-} & {-} & {-} & & {-} & {-} & {-} & {-} \\

      & Core-Set  \cite{core_set} & D & 23.10 & 16.73 & 14.19 & & 8.32 & 8.25 & 1.90 & 1.88 & & 34.11 & 46.25 & 18.59 & 32.41 \\

      & BADGE \cite{badge} & H & \underline{24.41} & \underline{17.55} & \underline{14.81} & & 8.35 & 8.33 & 1.74 & 1.74 & & 35.81 & 47.62 & 15.77 & 30.05 \\

      & DDFH \cite{ddfh} & H & 23.13 & 16.77 & 13.96 & & 8.22 & 8.15 & 1.85 & 1.84 & & 35.70 & 47.54 & 19.02 & 32.71 \\
      

      & CDAL \cite{cdal} & H & 23.30 & 16.69 & 14.04 & & 7.88 & 7.81 & 1.70 & 1.69 & & 35.25 & 47.20 & 15.43 & 29.87 \\

      \midrule

      \multirow{8}{*}{\rotatebox{90}{\textbf{Instance-based}}} 
      & Rand & - & 22.50 & 16.61 & 13.89 & & 9.28 & 9.20 & 2.20 & 2.18 & & 33.53 & 45.83 & 12.31 & 27.50 \\

      & Conf & U & 13.00 & 10.00 & 8.67 & & 9.05 & 8.97 & 2.32 & 2.31 & & 34.85 & 46.86 & 14.38 & 29.05 \\

      & Ens Depth Var & U & 11.29 & 9.01 & 8.43 & & 9.31 & 9.24 & 2.26 & 2.24 & & 34.28 & 46.38 & 16.68 & 30.95 \\

      & Augm Depth Var & U & 9.69 & 7.38 & 6.58 & & {-} & {-} & {-} & {-} & & {-} & {-} & {-} & {-} \\

      & ComPAS \cite{ComPAS} & U & 21.59 & 15.79 & 13.52 & & 9.25 & 9.18 & 2.24 & 2.23 & & 34.58 & 46.68 & 13.32 & 28.27 \\

      & Core-Set \cite{core_set} Box$_\text{3D}$ & D & 22.10 & 16.45 & 14.31 & & 9.85 & 9.78 & 2.38 & 2.37 & & 34.51 & 46.57 & \underline{20.08} & \underline{33.64} \\

      & BADGE \cite{badge} & H & 20.59 & 15.33 & 13.42 & & \underline{9.88} & \underline{9.80} & \underline{2.44} & \underline{2.42} & & \underline{36.13} & \underline{47.93} & 17.50 & 31.73 \\

      & \ourmethod ~{\scriptsize (Ours)}& D & \textbf{26.35} & \textbf{19.04} & \textbf{16.23} & & \textbf{10.18} & \textbf{10.10} & \textbf{2.60} & \textbf{2.59} & & \textbf{37.94} & \textbf{49.39} & \textbf{22.21} & \textbf{35.46} \\
      
      \bottomrule
    \end{tabularx}
    \label{tab:supp_main_results_final}
    \end{minipage}%
    \vspace{-0.8em}
\end{table*}

We provide additional comparisons to contextualize the main results in \cref{tab:main_results}. Final accuracies at the end of \ac{AL} training are summarized in \cref{tab:supp_main_results_final}, and the corresponding training curves are shown in \cref{fig:12_kitti_result_qualit,fig:12_ropescore_plot,fig:12_waymo_plots}, offering both endpoint and trajectory views of performance.

\begin{figure}
  \centering
   \includegraphics[width=0.49\textwidth]{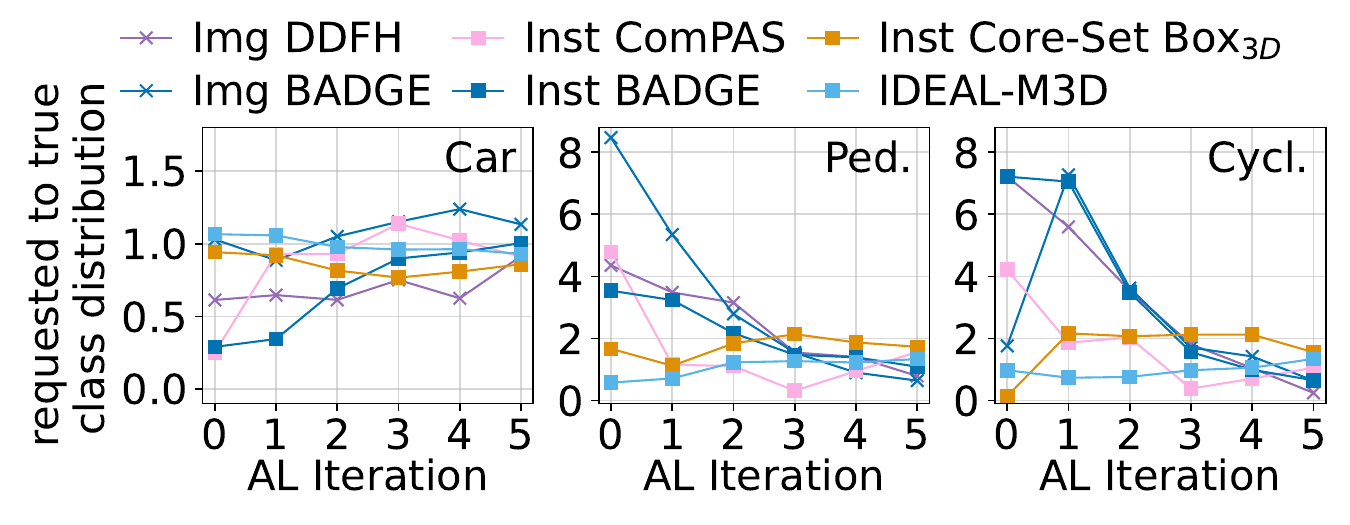}
  \caption{KITTI \cite{kitti} ratio of requested AL instances to the true dataset distribution ($<$1: Undersampling, $1$: Balanced, $>$1: Oversampling).} 
  \label{fig:cls_balance}
\end{figure}

Per-class results on KITTI for pedestrian and cyclist are reported in \cref{fig:12_kitti_result_qualit,tab:kitti_all_classes}. \ourmethod significantly outperforms all baselines on Car and Pedestrian. While ComPAS~\cite{ComPAS}, \coreSetBox BADGE~\cite{badge}, and DDFH~\cite{ddfh} report higher AP on Cyclist, this comes from a highly skewed budget allocation: they assign up to $7\times$ more annotations to Cyclist at the expense of the other classes (\cf \cref{fig:cls_balance}). In contrast, \ourmethod maintains a balanced acquisition across categories, which we hypothesize emerges from the representational diversity of our ensembles. This yields stronger average performance and a more uniform gain among all classes.

\subsection{Uncertainty vs. Diversity-based Methods}
\label{sec:div_vs_unc_methods}
\begin{figure}
  \centering
  \includegraphics[width=0.49\textwidth]{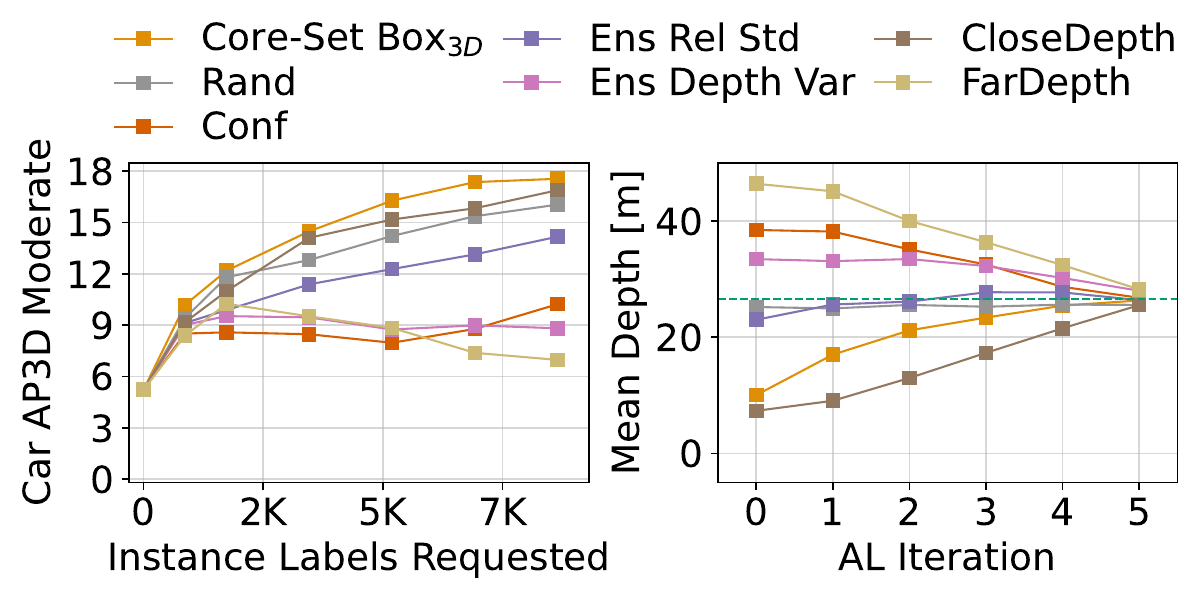}
  \caption{\textbf{Comparison of instance-based active learning methods for monocular 3D detection.} Uncertainty-based methods tend to select distant objects, which provide weaker training signals, while diversity-based approaches favor closer objects that are more informative. \textbf{Left}: The effectiveness of various selection strategies. \textbf{Right}: Distribution of selected instances, showing diversity-based methods' preference for close objects compared to uncertainty-based methods' bias toward moderately far and distant objects.}
  \label{fig:uncs_vs_div}
\end{figure}
To understand why uncertainty-based methods are less effective for instance-based \ac{M3D} (\cf \cref{tab:main_results}), we conduct a detailed analysis comparing \coreSetBox with three uncertainty-based methods:

\begin{itemize}
    \item \textit{Conf}: Selects instances with lowest model confidence scores (aleatoric uncertainty)
    \item \textit{Ens Rel Std}: Targets instances with highest relative depth deviation from the ensemble mean (three models)
    \item \textit{Ens Depth Var}: Prioritizes instances with highest absolute depth error relative to the ensemble mean
\end{itemize}

Our analysis of the selected instances in \cref{fig:uncs_vs_div}  reveals distinct selection patterns: Both \textit{Conf} and \textit{Ens Depth Var} demonstrate a clear bias towards distant objects, which is expected given that absolute depth errors typically increase with distance. In contrast, \textit{Ens Rel Std} achieves more balanced sampling across different depths, while \coreSetBox shows a preference for closer objects. We hypothesize that closer objects generally provide richer visual information due to higher pixel density and more distinct features, making them more conducive to accurate depth learning. Interestingly, diversity-based methods also tend to perform better for far away objects (hard category, \cf \cref{tab:main_results}), even though they are undersampled during training time. To further investigate this, we introduce two simple \ac{AL} strategies:
\begin{itemize}
    \item \textit{CloseDepth}: Prioritizes close instances for labeling
    \item \textit{FarDepth}: Prioritizes most distant instances (filtered to instances $\leq25$ pixels and $\leq50$m depth to avoid false positives)
\end{itemize}

Our experiments in \cref{fig:uncs_vs_div} show that \textit{CloseDepth} performs second-best after \coreSetBox, leading to two key insights. First, closer instances provide more effective training signals, even for detecting distant objects. Second, uncertainty-based methods are suboptimal for instance-based \ac{M3D} due to their inherent bias towards moderate far and distant objects

\subsection{Training time comparison with fully supervised methods}

We compare total training time in \cref{tab:training_time_wrt_supervised} against fully supervised and semi-supervised baselines. Relative to MonoLSS~\cite{monolss}, \ourmethod trains for roughly twice as long. It reaches (near) fully supervised accuracy using only 60\% of the labeled data on the validation and test sets. Since compute is typically far cheaper than human annotation, this is a favorable trade-off.

Compared to MonoLiG~\cite{monolig}, \ourmethod achieves higher accuracy on KITTI validation and test (\cf \cref{tab:kitti_test,fig:main_results_curves}). It also reduces training time by about $3\times$. This efficiency stems from concrete design choices. MonoLiG trains an ensemble of five image-based student models and an additional sixth LiDAR-based model. In contrast, we train only three models. Our auxiliary ensemble components are lightweight and fast to train.

We also compare with the semi-supervised Mix-Teaching~\cite{mix-teaching}. Its training time is more than $4\times$ higher. The method trains a five-model ensemble across one supervised and three semi-supervised rounds, which introduces substantial overhead. In contrast, \ourmethod attains higher KITTI test accuracy without using unlabeled data and with fewer labeled samples (\cf \cref{tab:kitti_test}).

\begin{table}[h]
  \centering
  \caption{Comparison of training time of selected methods on the KITTI \cite{kitti} trainval set. \textbf{SSL} denotes that the method uses semi-supervised learning. \textbf{AL} denotes that the method uses active learning.}
  \resizebox{\columnwidth}{!}{%
    \begin{tabular}{lccl}
      \textbf{Method} & SSL & AL & \textbf{Total training time} \\ \midrule
      MonoLSS \cite{monolss} \scriptsize{(Baseline)} & \xmark & \xmark & \cbarm{teal!65}{0.667em}{0.7em}{37h} \\
      \ourmethod (\scriptsize{Ours}) & \xmark & \cmark & \cbarm{teal!95}{1.352em}{0.7em}{75h} \\
      MonoLiG \cite{monolig} & \cmark & \cmark & \cbarm{teal!65}{4.32em}{0.7em}{240h} \\
        Mix-Teaching \cite{mix-teaching} & \cmark & \xmark & \cbarm{teal!65}{5.5em}{0.7em}{305h} \\
  \end{tabular}
  }
  \label{tab:training_time_wrt_supervised}
\end{table}

\subsection{Backbone ablation}
\begin{figure}[H]
  \centering
   \includegraphics[width=0.42\textwidth]{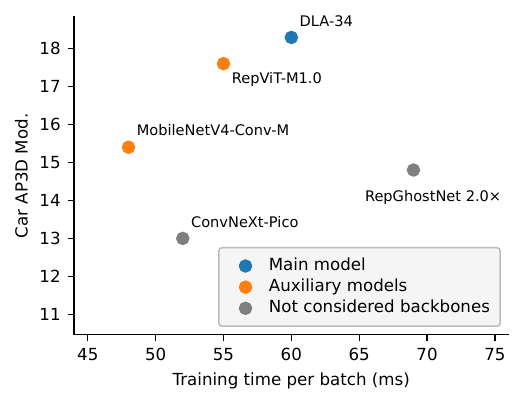}
  \caption{Comparing KITTI \cite{kitti} validation set performance under diverse backbones (supervised raining with 100\% of data).} 
  \label{fig:backbone_ablation}
\end{figure}

To promote ensemble diversity while keeping compute modest, we equip the auxiliary models with distinct lightweight backbones (cf.\ \cref{sec:diverse_ensembles}). We evaluate several candidates on KITTI~\cite{kitti} validation set using 100\% of the training data to isolate backbone effects. Throughput and accuracy are summarized in \cref{fig:backbone_ablation}.

For the main model, we retain DLA-34 \cite{dla} due to its strong accuracy, ensuring a fair and comparable reference across experiments. This backbone remains fixed in all primary results.

For the auxiliary models, we select RepViT-M1.0~\cite{repvit} and MobileNetV4-Conv-M~\cite{mobilenetv4}. MobileNetV4-Conv-M offers the best speed–accuracy trade-off among the tested lightweight backbones, being both faster and more accurate than alternatives. ConvNeXt-Pico~\cite{convnext} is marginally faster than RepViT-M1.0 but is approximately 25\% less accurate; we therefore prefer RepViT-M1.0. Together, RepViT-M1.0 and MobileNetV4-Conv-M provide complementary inductive biases and increase architectural diversity at low training cost.

\subsection{Feature diversity ablation}
\label{sec:analysis_feature_diversity}
\begin{table}
    \caption{\textbf{Ensemble ablation study on the KITTI \cite{kitti} validation set}. Results are reported using \ac{NAURC}$_{60\%}$ $AP_{3D|R_{40}}$ for cars (IoU=0.7) and pedestrians/cyclists (IoU=0.5). \textbf{Final AP}: Moderate AP after training on 60\% of the data.}
    \centering
    \small
    \begin{tabularx}{\linewidth}{
        @{}X 
        S[table-format=2.2]@{\hspace{0.2em}}
        S[table-format=2.2]@{\hspace{0.2em}}
        S[table-format=2.2]@{\hspace{0.2em}}
        S[table-format=2.2]@{\hspace{0.2em}}
        S[table-format=2.2]
        }
      \toprule
      Method& {Easy} & {Moderate} & {Hard} & {Final AP} \\ \midrule
      Ours w/o diverse backbones & 20.85 & 15.08  & 12.66  & 18.10 \\
      Ours w/o data sampling& 21.21 & 15.33  & 12.61 & 18.78 \\
      Ours w/ full epochs & 22.38 & 16.03  & 13.42  & 18.42 \\
      Ours w/o random loss & \ 22.02  & 15.55  & 12.55 & 18.93\\
      \midrule     
      \ourmethod (\scriptsize{Ours}) & \textbf{22.74} & \textbf{16.18} & \textbf{13.57} & {\textbf{19.04}} \\ 
      \bottomrule 
      \label{tab:ensemble_ablation}
      \end{tabularx}\vspace{-1.0em}
\end{table}

\begin{figure}
  \centering
  \begin{subfigure}[b]{0.155\textwidth}
    \includegraphics[width=\textwidth]{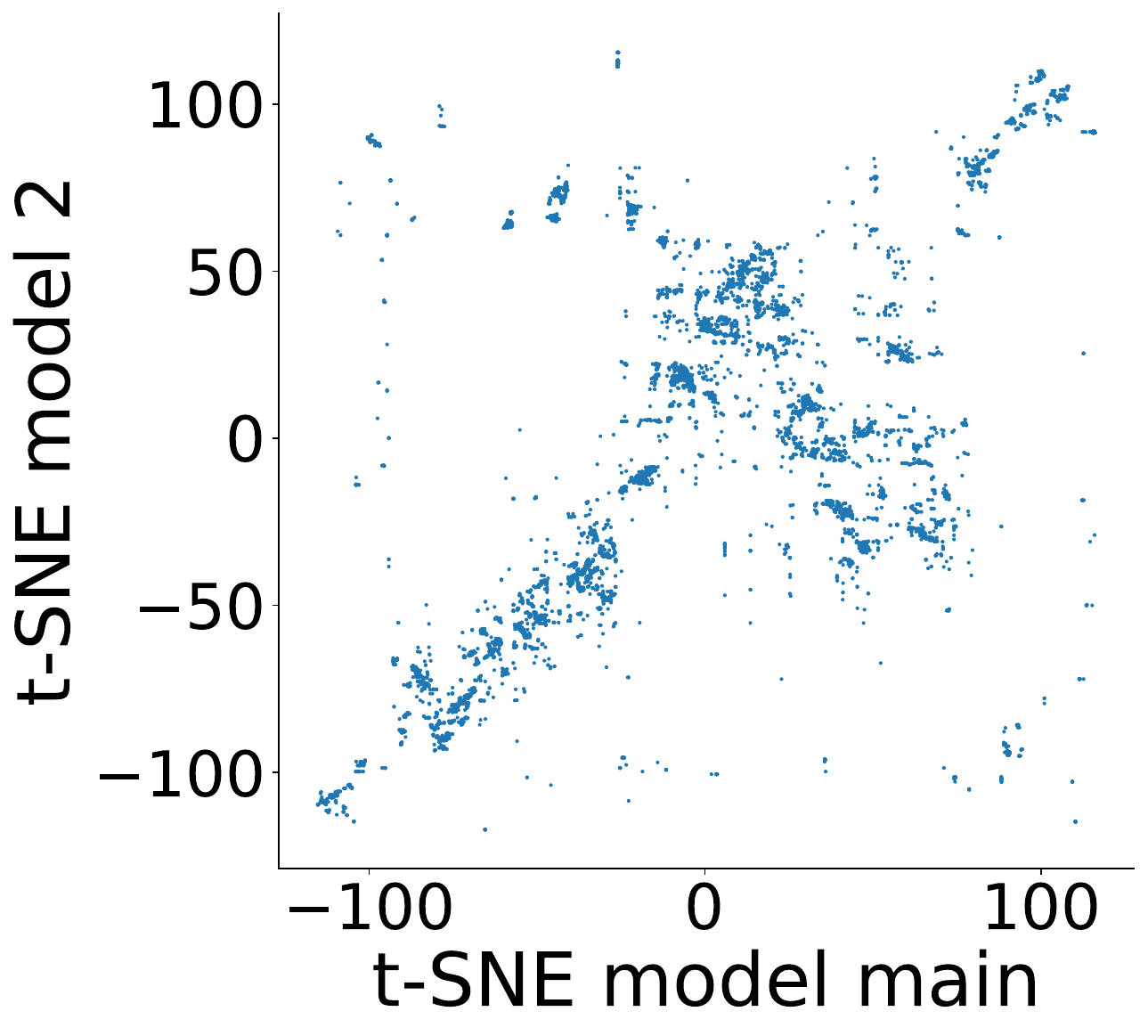}
    \caption{}\label{fig:04_feature_ensemble_main_main2}
  \end{subfigure}
  \begin{subfigure}[b]{0.155\textwidth}
    \includegraphics[width=\textwidth]{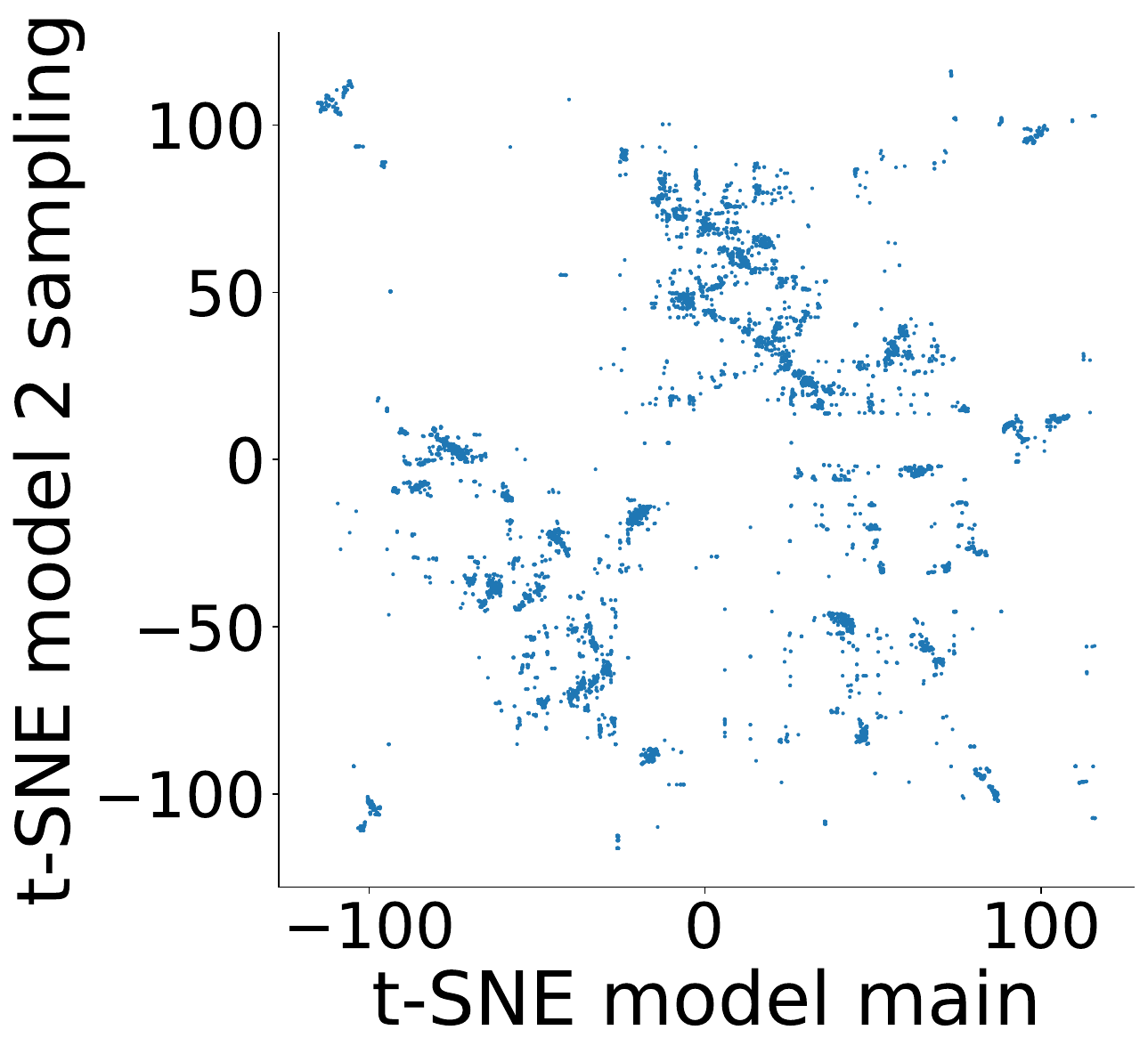}
    \caption{}\label{fig:04_feature_ensemble_main_sampling}
  \end{subfigure}
  \begin{subfigure}[b]{0.155\textwidth}
    \includegraphics[width=\textwidth]{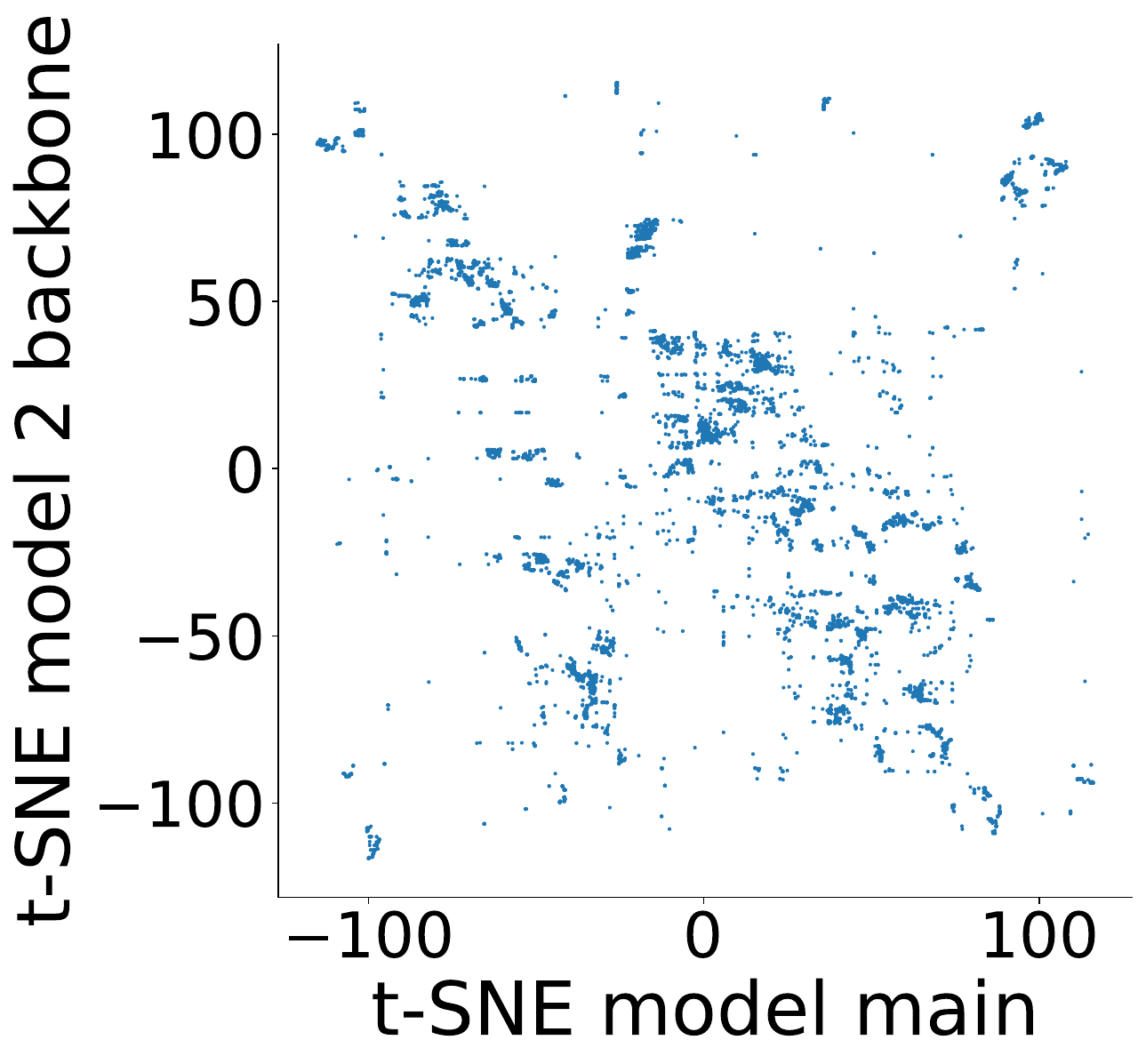}
    \caption{}\label{fig:04_feature_ensemble_main_backbone}
  \end{subfigure}
  \caption{\textbf{t-SNE \cite{tsne} visualization of \ac{RoI}}. Considered features from ensemble models trained on 30\% of the KITTI \cite{kitti} dataset.
\textbf{(a)} Identical data and backbones yield correlated features, limiting diversity.
\textbf{(b)-(c)} Adaptive sampling and varied backbones (RepViT \cite{repvit}) enhance feature diversity, improving ensemble effectiveness.} 
  \label{fig:05_tsne_features}
\end{figure}

A key component of our approach is instance selection driven by diverse, fast-to-train feature ensembles. We increase diversity and reduce compute through four complementary mechanisms: (i) heterogeneous lightweight backbones, (ii) time-adaptive data sampling that varies the training trajectory across ensemble members, (iii) fewer training epochs for auxiliary models, and (iv) random sampling of loss weights to perturb optimization and feature emphasis across tasks/heads.

Together, these mechanisms reduce total training time by approximately 15 hours compared to a vanilla, homogeneous ensemble trained for full schedules, while improving selection quality through more diverse feature views.

The ablation in \cref{tab:ensemble_ablation} supports this design. Reducing epochs on auxiliary models preserves performance to within noise levels, indicating that full schedules are unnecessary for effective feature-based selection. Adding backbone diversity, time-adaptive sampling, and random loss-weight sampling yields incremental gains in AP$_{3D|R_{40}}$ (Moderate) of +1.10, +0.85, and +0.63, respectively.

We further analyze feature diversity in \cref{fig:05_tsne_features} via t-SNE~\cite{tsne}. Ensembles using time-adaptive sampling and heterogeneous backbones exhibit markedly lower inter-model feature correlation than a vanilla ensemble, confirming that our mechanisms produce complementary representations that enhance the quality of selected instances.

\subsection{Visual diversity ablation}
\begin{figure}
  \centering
  \begin{subfigure}[b]{0.155\textwidth}
    \vspace{-1em}
    \includegraphics[width=\textwidth]{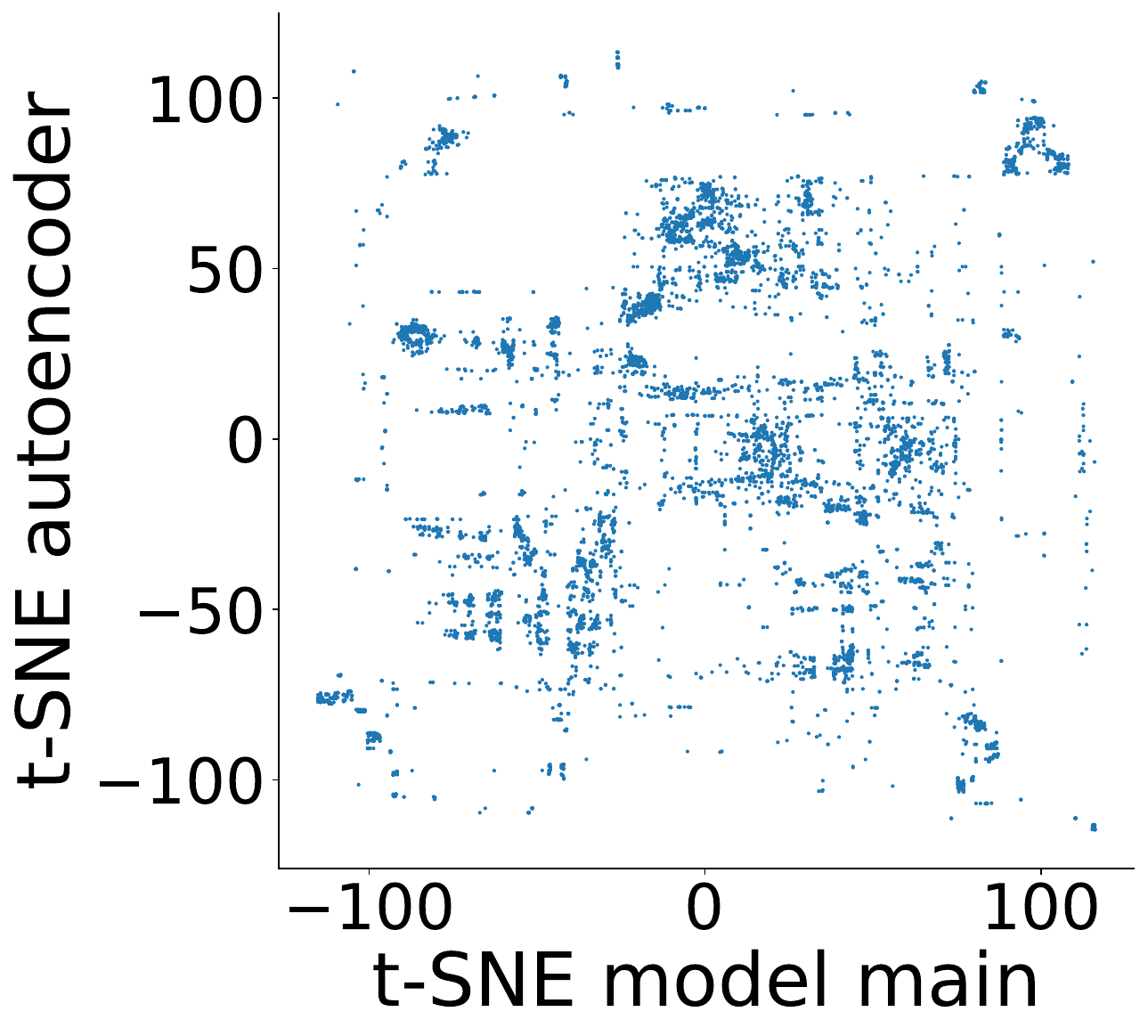}

    \caption{}\label{fig:04_method_features_main_autoencoder}
  \end{subfigure}
  \begin{subfigure}[b]{0.31\textwidth}
    \includegraphics[width=\textwidth]{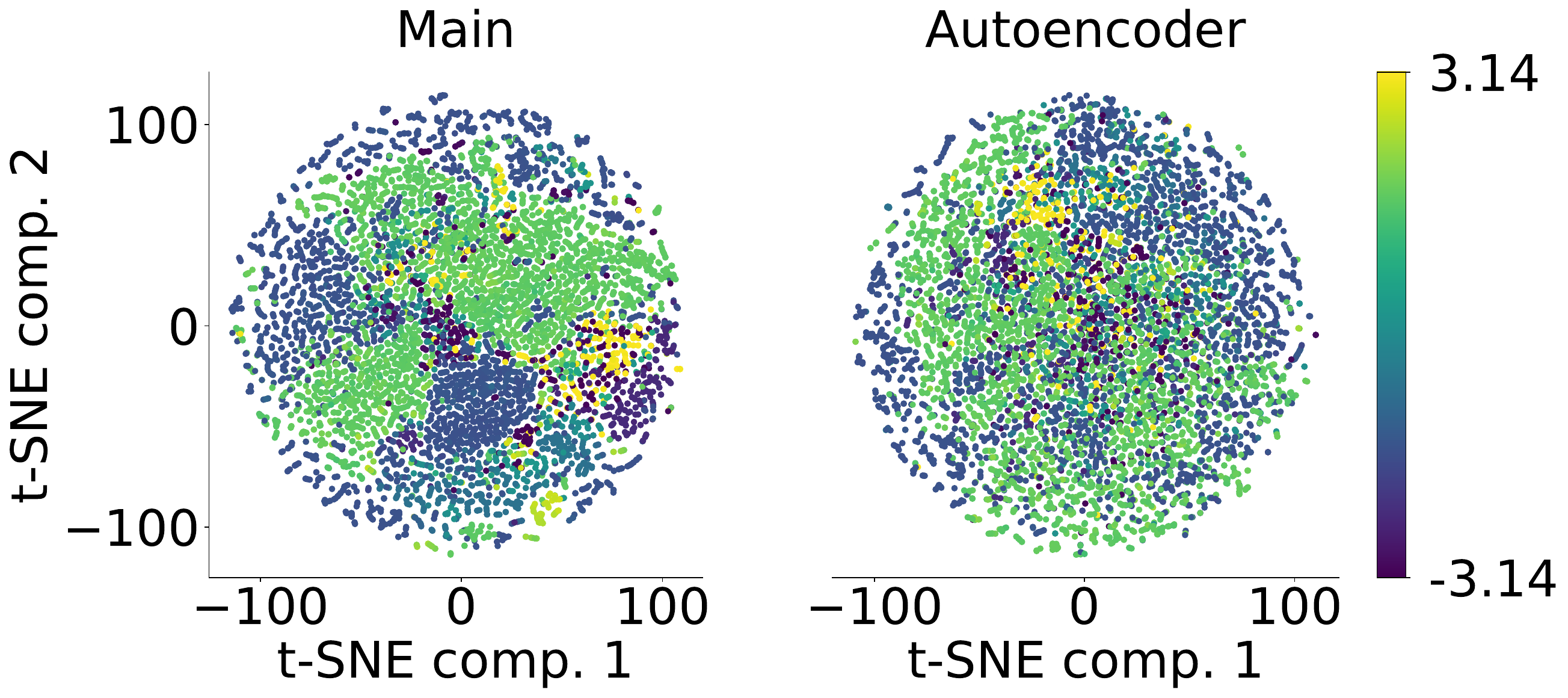}
    \caption{}\label{fig:04_method_features_main_angle}
  \end{subfigure}
  \caption{\textbf{t-SNE \cite{tsne} visualization of features trained on 30\% of the KITTI \cite{kitti} dataset, highlighting the complementarity of \ac{RoI} and autoencoder features.}
\textbf{(a)} \ac{RoI} and autoencoder features show low correlation, capturing distinct information.
\textbf{(b)} Example: \ac{RoI} features are dominated by the object orientation around the $y$-axis, while Stable Diffusion v2-base \cite{ldm} autoencoder features are more orientation-invariant. The vehicle's orientation is shown by color. } 
  \label{fig:tsne_features_ldm}
\end{figure}

\begin{table}
    \caption{\textbf{Visual diversity ablation study on the KITTI \cite{kitti} validation set}. Results are reported using \ac{NAURC}$_{60\%}$ $AP_{3D|R_{40}}$ for cars (IoU=0.7) and pedestrians/cyclists (IoU=0.5). \textbf{Final AP}: Moderate AP after training on 60\% of the data.}
    \centering
    \small
    \begin{tabularx}{\linewidth}{
        @{}X 
        S[table-format=2.2]@{\hspace{0.2em}}
        S[table-format=2.2]@{\hspace{0.2em}}
        S[table-format=2.2]@{\hspace{0.2em}}
        S[table-format=2.2]@{\hspace{0.2em}}
        S[table-format=2.2]
        }
      \toprule
      Method& {Easy} & {Moderate} & {Hard} & {Final AP} \\ \midrule
      Ours w/o SAMv2 \cite{sam} & 17.72 & 13.24  & 11.11 & 16.36 \\
      Ours w/ DINOv2 \cite{dinov2} & 17.87 & 13.08  & 11.12 & 16.60 \\
      \midrule     
      \ourmethod (\scriptsize{Ours}) & \textbf{22.74} & \textbf{16.18} & \textbf{13.57} & {\textbf{19.04}} \\ 
      \bottomrule 
      \label{tab:visual_diversity_ablation}
      \end{tabularx}\vspace{-1.0em}
\end{table}

In \cref{tab:visual_diversity_ablation}, we ablate components that promote visual diversity in the feature ensemble.

Removing \ac{SAMv2} reduces $AP_{3D}$ (Moderate) by more than 2 points. We hypothesize two causes. First, without \ac{SAMv2}, background content leaks into the features. The same object with different backgrounds then appears artificially diverse, which dilutes foreground cues. Second, the foreground mask carries coarse geometric information. E.g, the silhouette of a car at $45^\circ$ yaw differs from one at $0^\circ$, which is useful for 3D selection.

Replacing the autoencoder with DINOv2~\cite{dinov2} features causes a similar drop. DINOv2 emphasizes global scene semantics. It captures less local, instance-centric detail~\cite{ohaDino}. The autoencoder yields object-focused representations that better support instance selection.

\cref{fig:tsne_features_ldm} shows t-SNE~\cite{tsne} visualizations of autoencoder features and detector features. The two spaces exhibit minimal correlation. This indicates that the task-agnostic autoencoder provides complementary signals to the task-specific model features.

\label{sec:div_ensemble}

\subsection{Qualitative Results}
\begin{figure*}
  \centering
   \includegraphics[width=0.98\textwidth]{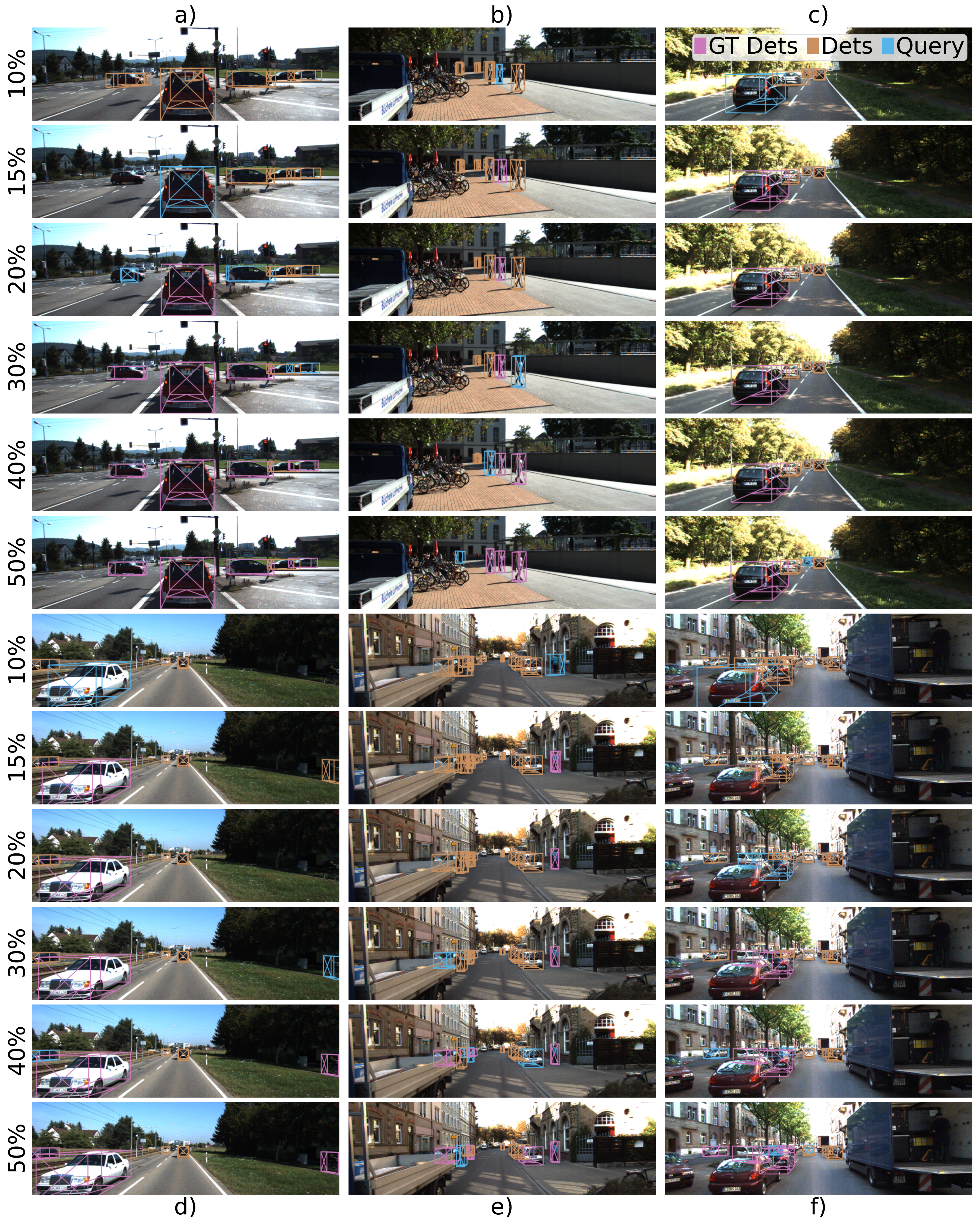}

  \caption{\textbf{Qualitative results of \ac{IDEAL-M3D} on the KITTI ~\cite{kitti} dataset showing prediction evolution over time (top to bottom)}. Pink boxes represent previously labeled instances now in the training set, cyan boxes indicate predictions selected for current labeling, and orange boxes show predictions not chosen for annotation. The progression demonstrates the model's improvement through strategic label acquisition
  }
  \label{fig:qual_results_kitti}
\end{figure*}

\begin{figure*}
  \centering
   \includegraphics[width=0.7\textwidth]{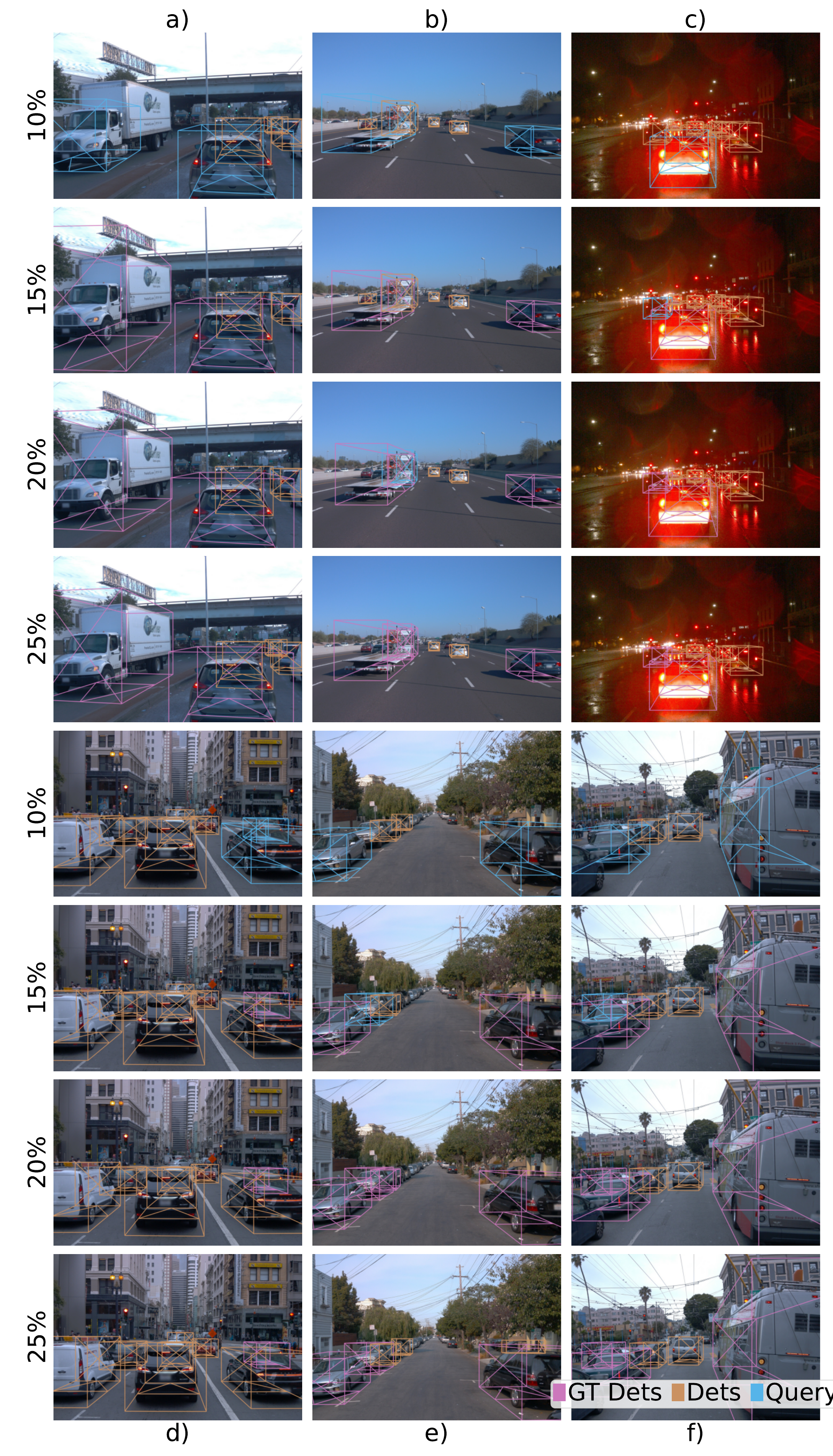}

  \caption{\textbf{Qualitative results of \ac{IDEAL-M3D} on the Waymo ~\cite{Waymo} dataset showing prediction evolution over time (top to bottom)}. Pink boxes represent previously labeled instances now in the training set, cyan boxes indicate predictions selected for current labeling, and orange boxes show predictions not chosen for annotation. The progression demonstrates the model's improvement through strategic label acquisition
  }
  \label{fig:qual_results_waymo}
\end{figure*}

\begin{figure*}
  \centering
   \includegraphics[width=0.8\textwidth]{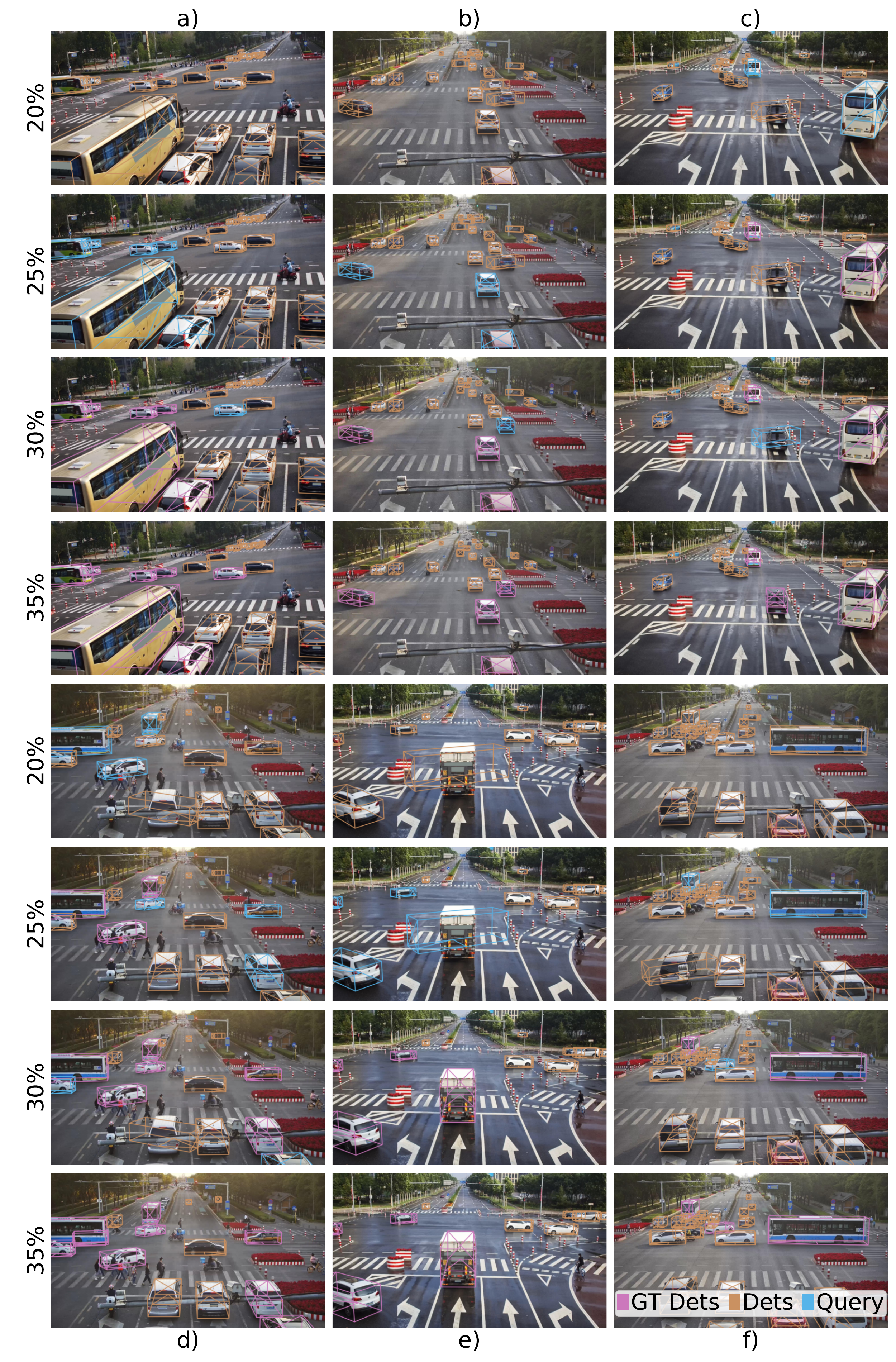}

  \caption{\textbf{Qualitative results of \ac{IDEAL-M3D} on the Rope3D ~\cite{rope3d} dataset showing prediction evolution over time (top to bottom)}. Pink boxes represent previously labeled instances now in the training set, cyan boxes indicate predictions selected for current labeling, and orange boxes show predictions not chosen for annotation. The progression demonstrates the model's improvement through strategic label acquisition
  }
  \label{fig:qual_results_rope3d}
\end{figure*}

In the subsequent pages (\cf \cref{fig:qual_results_kitti}, \cref{fig:qual_results_waymo}, \cref{fig:qual_results_rope3d}) we present further qualitative results on \ac{IDEAL-M3D} highlighting the prediction accuracy and selection process over time. 

\end{document}